\documentclass[10pt,journal,compsoc]{IEEEtran}
\ifCLASSOPTIONcompsoc
  \usepackage[nocompress]{cite}
\else
  \usepackage{cite}
\fi
\ifCLASSINFOpdf
\else
\fi

\usepackage{epsfig}
\usepackage{graphicx}
\usepackage{amsmath}
\usepackage{amssymb}
\usepackage{enumitem}

\usepackage[pagebackref=true,breaklinks=true,letterpaper=true,colorlinks,bookmarks=false,urlcolor=red,citecolor=cyan]{hyperref}
\usepackage{gensymb,graphics,subfigure,inputenc}
\usepackage[linesnumbered,algo2e,boxed]{algorithm2e}
\graphicspath{{Figure/}}
\usepackage[figuresright]{rotating}
\usepackage{amsmath,amssymb,textcomp,subfigure,multirow,upgreek}
\usepackage{booktabs}
\usepackage{color,mdwlist}

\newcommand{\hao}[1]{\textcolor{black}{#1}}
\usepackage{ragged2e} 

\usepackage{graphicx,fontawesome}
\graphicspath{{Figures/}}

\usepackage{caption}
\usepackage{enumitem}
\setitemize{noitemsep,topsep=0pt,parsep=0pt,partopsep=0pt}

\begin{document}
\title{Edge Guided GANs with Multi-Scale Contrastive Learning for Semantic Image Synthesis}
\author{Hao~Tang, Guolei Sun, Nicu Sebe, Luc Van Gool
	\IEEEcompsocitemizethanks{
	    \IEEEcompsocthanksitem Hao Tang, Guolei Sun, Luc Van Gool are with the Department of Information Technology and Electrical Engineering, ETH Zurich,  Zurich 8092, Switzerland. E-mail: hao.tang@vision.ee.ethz.ch \protect
        \IEEEcompsocthanksitem Nicu Sebe is with the Department of Information Engineering and Computer Science (DISI), University of Trento, Trento 38123, Italy. \protect
        }% <-this % stops an unwanted space
    	\thanks{Corresponding author: Hao Tang.}
}


\markboth{IEEE Transactions on Pattern Analysis and Machine Intelligence}%
{Shell \MakeLowercase{\textit{et al.}}: Bare Demo of IEEEtran.cls for Computer Society Journals}



\IEEEtitleabstractindextext{%
%\begin{abstract}
%The abstract goes here.
%\end{abstract}
\justify
\begin{abstract}
We propose a novel \underline{e}dge guided \underline{g}enerative \underline{a}dversarial \underline{n}etwork with \underline{c}ontrastive learning (ECGAN) for the challenging semantic image synthesis task. Although considerable improvements have been achieved by the community in the recent period, the quality of synthesized images is far from satisfactory due to three largely unresolved challenges. 1) The semantic labels do not provide detailed structural information, making it challenging to synthesize local details and structures; 2) The widely adopted CNN operations such as convolution, down-sampling, and normalization usually cause spatial resolution loss and thus cannot fully preserve the original semantic information, leading to semantically inconsistent results (e.g., missing small objects); 3) Existing semantic image synthesis methods focus on modeling ``local'' semantic information from a single input semantic layout. However, they ignore ``global'' semantic information of multiple input semantic layouts, i.e., semantic cross-relations between pixels across different input layouts. To tackle 1), we propose to use the edge as an intermediate representation which is further adopted to guide image generation via a proposed attention guided edge transfer module. Edge information is produced by a convolutional generator and introduces detailed structure information. To tackle 2), we design an effective module to selectively highlight class-dependent feature maps according to the original semantic layout to preserve the semantic information. To tackle 3), inspired by current methods in contrastive learning, we propose a novel contrastive learning method, which aims to enforce pixel embeddings belonging to the same semantic class to generate more similar image content than those from different classes.  We further propose a novel multi-scale contrastive learning method that aims to push same-class features from different scales closer together being able to capture more semantic relations by explicitly exploring the structures of labeled pixels from multiple input semantic layouts from different scales. Experiments on three challenging datasets show that our methods achieve significantly better results than state-of-the-art approaches. The source code is available at
\url{https://github.com/Ha0Tang/ECGAN}.
\end{abstract}

\begin{IEEEkeywords}
GANs, Edge Guided, Multi-Scale, Contrastive Learning, Semantic Image Synthesis. 

% Multi-scale and Cross-scale Contrastive Learning for Semantic Segmentation; 
% Semantic-shape Adaptive Feature Modulation for Semantic Image Synthesis; 
% Pretraining is All You Need for Image-to-Image Translation
\end{IEEEkeywords}}

\maketitle

\IEEEdisplaynontitleabstractindextext

% \fi
\IEEEpeerreviewmaketitle



\section{Introduction}

\begin{figure*} [!t] \small
	\centering
	\includegraphics[width=0.9\linewidth]{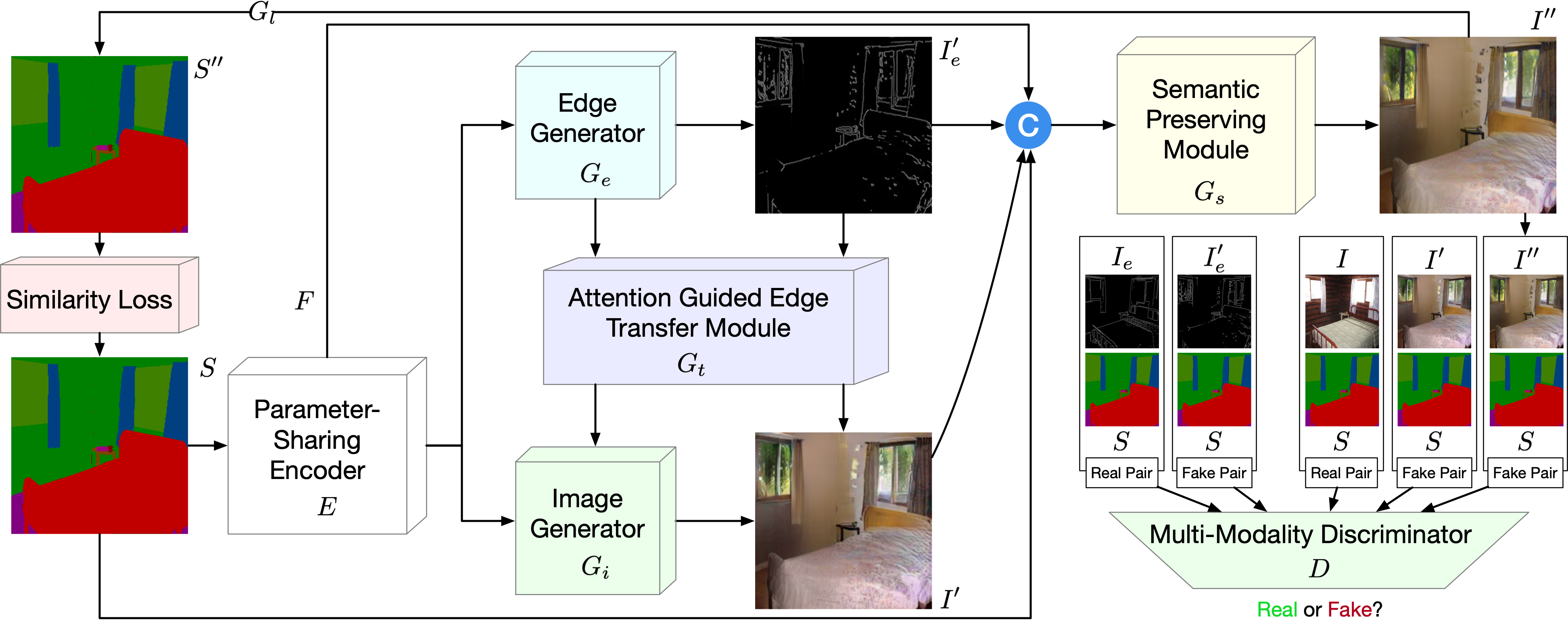}
	\caption{Overview of the proposed ECGAN. 
     It consists of a parameter-sharing encoder $E$, an edge generator $G_e$, an image generator $G_i$, an attention guided edge transfer module $G_t$, a label generator $G_l$, a similarity loss module, a contrastive learning module $G_c$ (not shown for brevity), and a multi-modality discriminator $D$. 
	 Both $G_e$ and $G_i$ are connected by $G_t$ from two levels, i.e., edge feature-level and content-level, to generate realistic images. $G_s$ is proposed to preserve the semantic information of the input semantic labels. 
	 $G_l$ aims to transfer the generated image back to the label for calculating the similarity loss. $G_c$ tries to capture more semantic relations by explicitly exploring the structures of labeled pixels from multiple input semantic layouts. $D$ aims to distinguish the outputs from two modalities, i.e., edge and image. The whole framework can be end-to-end trained so that each component can benefit from each other. The symbol ${\footnotesize \textcircled{c}}$ denotes channel-wise concatenation.
 }
	\label{fig:method}
	\vspace{-0.4cm}
\end{figure*}

Semantic image synthesis refers to generating photo-realistic images conditioned on pixel-level semantic labels.
This task has a wide range of applications such as image editing and content generation~\cite{chen2017photographic,isola2017image,gu2019mask,liu2019learning,qi2018semi}.
Although existing methods conducted interesting explorations, we still observe unsatisfactory aspects, mainly in the generated local structures and details, as well as small-scale objects, which we believe are mainly due to three reasons:
1) Conventional methods~\cite{park2019semantic,wang2018high,liu2019learning} generally take the semantic label map as input directly. 
However, the input label map provides only structural information between different semantic-class regions and does not contain any structural information within each semantic-class region, making it difficult to synthesize rich local structures within each class.
Taking label map $S$ in Figure~\ref{fig:method} as an example, the generator does not have enough structural guidance to produce a realistic bed, window, and curtain from only the input label ($S$).  
2) The classic deep network architectures are constructed by stacking convolutional, down-sampling, normalization, non-linearity, and up-sampling layers, which will cause the problem of spatial resolution losses of the input semantic labels.
3) Existing methods for this task are typically based on global image-level generation.
In other words, they accept a semantic layout containing several object classes and aim to generate the appearance of each one using the same network. In this way, all the classes are treated equally. 
However, because different semantic classes have distinct properties, using specified network learning for each would intuitively facilitate the complex generation of multiple classes.

To address these three issues, in this paper, we propose a novel \underline{e}dge guided \underline{g}enerative \underline{a}dversarial \underline{n}etwork with \underline{c}ontrastive learning (ECGAN) for semantic image synthesis. 
The overall framework of ECGAN is shown in Figure~\ref{fig:method}. To tackle 1), we first propose an edge generator to produce the edge features and edge maps. 
Then the generated edge features and edge maps are selectively transferred to the image generator and improve the quality of the synthesized image by using our attention guided edge transfer module. To tackle 2), we propose an effective semantic preserving module, which aims at selectively highlighting class-dependent feature maps according to the original semantic layout. 
We also propose a new similarity loss to model the relationship between semantic categories. Specifically, given a generated label $S''$ and corresponding ground truth $S$, similarity loss constructs a similarity map to supervise the learning. To tackle 3), a straightforward solution would be to model the generation of different image classes individually. By so doing, each class could have its own generation network structure or parameters, thus greatly avoiding the learning of a biased generation space. 
However, there is a fatal disadvantage to this. That is, the number of parameters of the network will increase linearly with the number of semantic classes $N$, which will cause memory overflow and make it impossible to train the model.
If we use $p_e$ and $p_d$ to denote the number of parameters of the encoder and decoder, respectively, then the total number of the network parameter should be $p_e {+} N{\times} p_d$ since we need a new decoder for each class.
To further address this limitation, we introduce a pixel-wise contrastive learning approach that elevates the current image-wise training method to a pixel-wise method. By leveraging the global semantic similarities present in labeled training layouts, this method leads to the development of a well-structured feature space.
In this case, the total number of the network parameter only is $p_e {+} p_d$.
Moreover, we explore image generation from a class-specific context, which is beneficial for generating richer details compared to the existing image-level generation methods. A new class-specific pixel generation strategy is proposed for this purpose. It can effectively handle the generation of small objects and details, which are common difficulties encountered by the global-based generation.

With the proposed ECGAN, we achieve new state-of-the-art results on Cityscapes~\cite{cordts2016cityscapes}, ADE20K~\cite{zhou2017scene}, and COCO-Stuff \cite{caesar2018coco} datasets, demonstrating the effectiveness of our approach in generating images with complex scenes and showing significantly better results compared with existing methods.
To summarize, our contributions are as follows:

\begin{itemize}
	\item We propose a novel ECGAN for the challenging semantic image synthesis task. To the best of our knowledge, we are the first to explore the edge generation from semantic layouts and then utilize the generated edges to guide the generation of realistic images.  
    \item  We propose an effective attention guided edge transfer module to selectively transfer useful edge structure information from the edge generation branch to the image generation branch.
    \item  We design a new semantic preserving module to highlight class-dependent feature maps based on the input semantic label map for generating semantically consistent results. 
    \item We propose a new similarity loss to capture the intra-class and inter-class semantic dependencies, leading to robust training.
    \item We propose a novel contrastive learning method, which learns a well-structured pixel semantic embedding space by utilizing global semantic similarities among labeled layouts. Moreover, we propose a multi-scale contrastive learning method with two novel multi-scale and cross-scale losses that enforces local-global feature consistency between low-resolution global and high-resolution local features extracted from different scales.
    \item We conduct extensive experiments on three challenging datasets under diverse scenarios, i.e., Cityscapes~\cite{cordts2016cityscapes}, ADE20K~\cite{zhou2017scene}, and COCO-Stuff~\cite{caesar2018coco}. Both qualitative and quantitative results show that the proposed methods are able to produce remarkably better results than existing baseline models regarding both visual fidelity and alignment with the input semantic layouts. Moreover, our methods can generate multi-modal images and edges, which have not been considered by existing state-of-the-art methods. 
\end{itemize}

Part of the material presented here appeared in \cite{tang2023edge}. The current paper extends \cite{tang2023edge} in several ways.
(1) We present a more detailed analysis of related works by including recently published works dealing with semantic image synthesis and contrastive learning.
(2) We propose a novel module, i.e., multi-scale contrastive learning, to push the same-class features from different scales to be similar by using the proposed multi-scale and cross-scale contrastive learning losses.
Equipped with this new module, our ECGAN proposed in \cite{tang2023edge} is upgraded to ECGAN++.
(3) We extend the quantitative and qualitative experiments by comparing our ECGAN and ECGAN++ with the very recent works on three public datasets. Extensive experiments show that the proposed ECGAN++ achieves the best results compared with existing methods.

\section{Related Work}

\noindent \textbf{Generative Adversarial Networks (GANs)} \cite{goodfellow2014generative} have two important components, i.e., a generator and a discriminator. Both are trained in an adversarial way to achieve a balance. 
Recently, GANs have shown the capability of generating realistic images \cite{tang2021total,tang2020unified}.
Moreover, to generate user-specific content, Conditional GANs (CGANs) \cite{mirza2014conditional} have been proposed.
CGANs usually combine a vanilla GAN and some external information such as class labels \cite{tang2019attribute,tang2019expression}, human poses \cite{tang2022bipartite,tang2022facial,tang2020xinggan,tang2019cycle,tang2018gesturegan}, text descriptions \cite{xu2022predict,tao2022df,tao2023galip,tao2022net}, graphs \cite{tang2023graph}, and segmentation maps \cite{gu2019mask,wu2022cross,tang2022local,tang2020dual,tang2020local,wu2022cross_tmm,tang2021layout}.

\noindent \textbf{Image-to-Image Translation} aims to generate the target image based on an input image. 
CGANs have achieved decent results in both paired \cite{isola2017image,albahar2019guided} and unpaired \cite{zhu2017unpaired} image translation tasks.
For instance, Isola et al. propose Pix2pix~\cite{isola2017image}, which employs a CGAN to learn a translation mapping from input to output image domains such as map-to-photo and day-to-night. 
Moreover, Zhu et al. \cite{zhu2017unpaired} introduce CycleGAN, which targets unpaired image-to-image translation using the cycle-consistency loss.
To further improve the quality of the generated images, the attention mechanism has been recently investigated in image translation tasks \cite{tang2019multi,tang2019attention,mejjati2018unsupervised,chen2018attention,tang2021attentiongan}. 
Attention mechanism assigns context elements weights which define a weighted sum over context representation \cite{wu2019pay,chen2019graph}, which has been used in many other computer vision tasks such as depth estimation \cite{xu2018structured} and semantic segmentation \cite{fu2019dual}, and have shown great effectiveness. 

Different from previous attention-related image generation works, we propose a novel attention guided edge transfer module to transfer useful edge structure information from the edge generation branch to the image generation branch at two different levels, i.e., feature level and content level.
To the best of our knowledge, our module is the first attempt to incorporate both edge feature attention and edge content attention within a GAN framework for image-to-image translation tasks.

\noindent \textbf{Edge Guided Image Generation.}
Edge maps are usually adopted in image inpainting \cite{ren2019structureflow,nazeri2019edgeconnect,li2019progressive} and image super-resolution \cite{nazeri2019edge} tasks to reconstruct the missing structure information of the inputs.
For example, 
Pix2pix~\cite{isola2017image} adopts edge maps as input and aims to generate realistic shoes and handbags, which can be seen as an edge-to-image translation problem.
\hao{Moreover, \cite{nazeri2019edgeconnect} proposed an edge generator to hallucinate edges in the missing regions given edges, which can be regarded as an edge completion problem. 
Using edge images as the structural guidance, EdgeConnect \cite{nazeri2019edgeconnect} achieves good results even for some highly structured scenes.
To recover meaningful structures, \cite{ren2019structureflow} implemented edge-preserved smooth images, serving as representations of the overarching structures inherent in image scenes. When these images are used as a navigational tool for the structure reconstructor, the network has the capacity to concentrate on the recuperation of these global structures, undeterred by any extraneous texture data.}

Unlike previous works, including \cite{nazeri2019edgeconnect,ren2019structureflow}, we propose a novel edge generator to perform a new task, i.e., semantic label-to-edge translation. To the best of our knowledge, we are the first to generate edge maps from semantic labels. Then the generated edge maps, with more local structure information, can be used to improve the quality of the image results.

\noindent \textbf{Semantic Image Synthesis} aims to generate a photo-realistic image from a semantic label map \cite{chen2017photographic,qi2018semi,park2019semantic,liu2019learning,bansal2019shapes,zhu2020sean,ntavelis2020sesame,zhu2020semantically,sushko2020you,tan2021efficient,tan2021diverse,zhu2020semantically,zhang2023adding,zeng2023scenecomposer,shi2022semanticstylegan}.
With semantic information as guidance, existing methods have achieved promising performance.
However, we can still observe unsatisfying aspects, especially on the generation of the small-scale objects, which we believe is mainly due to the problem of spatial resolution losses associated with deep network operations such as convolution, normalization, down-sampling, etc.
To solve this problem, \cite{park2019semantic}~proposed GauGAN, which uses the input semantic labels to modulate the activations in normalization layers through a spatially-adaptive transformation.
However, the spatial resolution losses caused by other operations, such as convolution and down-sampling, have not been resolved.
Moreover, we observe that the input label map has only a few semantic classes in the entire dataset. 
Thus the generator should focus more on learning these existing semantic classes rather than all the semantic classes.

To tackle both limitations, we propose a novel semantic preserving module, which aims to selectively highlight class-dependent feature maps according to the input labels for generating semantically consistent images. 
We also propose a new similarity loss to model the intra-class and inter-class semantic dependencies.

\begin{figure*} [t]
	\centering
	\includegraphics[width=0.85\linewidth]{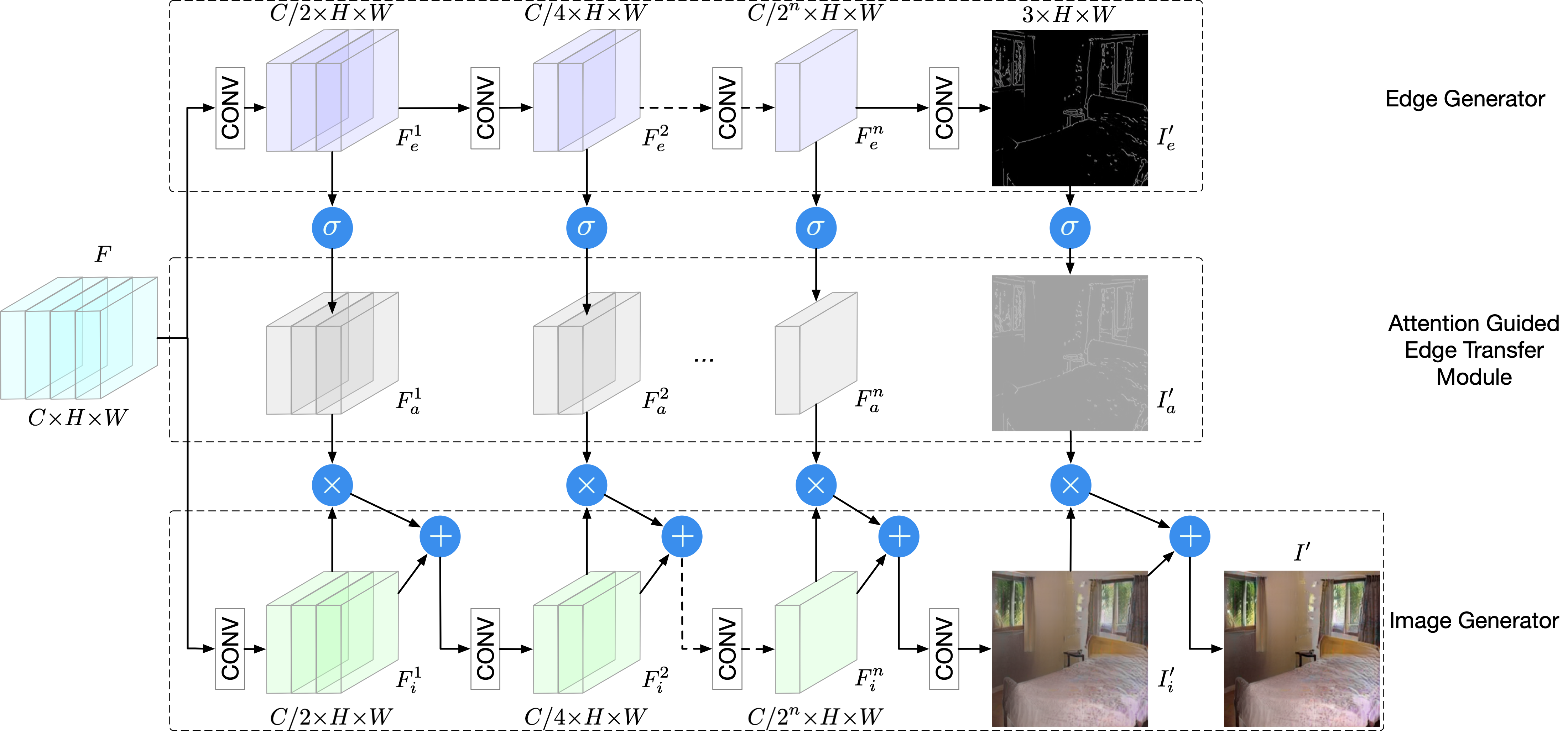}
	\caption{Structure of the proposed edge generator $G_e$, attention guided edge transfer module $G_t$, and image generator $G_i$. $G_e$ selectively transfers useful local structure information to $G_i$ using the proposed attention guided transfer module $G_t$. 
    The symbols $\oplus$, $\otimes$, and ${\footnotesize \textcircled{$\sigma$}}$ denote element-wise addition, element-wise multiplication, and Sigmoid activation function, respectively.}
	\label{fig:edge_block}
	\vspace{-0.4cm}
\end{figure*}

\noindent \textbf{Contrastive Learning.}
Recently, the most compelling
methods for learning representations without labels have
been unsupervised contrastive learning \cite{van2018representation,hjelm2018learning,pan2021videomoco,wu2018unsupervised,chen2020simple,zhao2021contrastive}, which significantly outperform other pretext task-based alternatives \cite{gidaris2018unsupervised,doersch2015unsupervised,noroozi2016unsupervised}. 
Contrastive learning aims to learn the general features of unlabeled data by teaching and guiding the model which data points are different or similar.
For example, 
\cite{pan2021videomoco} proposed VideoMoCo for unsupervised video representation learning.
\cite{chen2020simple} introduced a simple framework for contrastive learning of visual representations, which we call SimCLR.
\cite{hu2021region} designed a region-aware contrastive learning to explore semantic relations for the specific semantic segmentation problem.
Both \cite{wang2021exploring} and \cite{zhao2021contrastive} also proposed two new contrastive learning-based strategies for semantic segmentation.

However, we propose a novel contrastive learning method for semantic image synthesis in this paper. 
This synthesis task is very different from the semantic segmentation task, which requires us to tailor the network structure and loss function.
Specifically, we propose a new training protocol that explores global pixel relations in labeled layouts for regularizing the generation embedding space. 
Moreover, we extend it to a multi-scale version that can enforce local-global feature consistency between low-resolution global and high-resolution local features by introducing two new multi-scale and cross-scale contrastive learning losses.

\section{Edge Guided GANs with Contrastive Learning}

\noindent \textbf{Framework Overview.}
Figure~\ref{fig:method} shows the overall structure of ECGAN for semantic image synthesis, which consists of a semantic and edge guided generator $G$ and a multi-modality discriminator $D$.
The generator $G$ consists of eight components:
(1) a parameter-sharing convolutional encoder $E$ is proposed to produce deep feature maps $F$; 
(2) an edge generator $G_e$ is adopted to generate edge maps $I'_e$ taking as input deep features from the encoder;
(3) an image generator $G_i$ is used  to produce intermediate images $I'$;
(4) an attention guided edge transfer module $G_t$ is designed to forward useful structure information from the edge generator to the image generator;
(5) the semantic preserving module $G_s$ is developed to selectively highlight class-dependent feature maps according to the input label for generating semantically consistent images $I''$;
(6) a label generator $G_l$ is employed to produce the label from $I''$;
(7) the similarity loss is proposed to calculate the intra-class and inter-class relationships.
(8) the contrastive learning module $G_c$ aims to 
model global semantic relations between training pixels, guiding pixel embeddings towards cross-image
category-discriminative representations that eventually improve the generation performance.

Meanwhile, to effectively train the network, we propose a multi-modality discriminator $D$ that distinguishes the outputs from both modalities, i.e., edge and image.

\subsection{Edge Guided Semantic Image Synthesis}
\noindent \textbf{Parameter-Sharing Encoder.}
The backbone encoder $E$ can employ any deep network architecture, e.g., the commonly used AlexNet \cite{krizhevsky2012imagenet},
VGG \cite{simonyan2015very}, and ResNet \cite{he2016deep}. 
We directly utilize the feature maps  from the last convolutional layer as deep feature representations, i.e., $F {=} E(S)$, where $E$ represents the encoder; $S{\in} \mathbb{R}^{N \times H \times W}$ is the input label, with $H$ and $W$ as width and height of the input semantic labels, and $N$ as the total number of semantic classes.
Optionally, one can always combine multiple intermediate feature maps to enhance the feature representation.
The encoder is shared by the edge generator and the image generator.
Then, the gradients from the two generators all contribute  to updating the parameters of the encoder.
This compact design can potentially enhance the deep representations as the encoder can simultaneously learn structure representations from the edge generation branch and appearance representations from the image generation branch.

\begin{figure*}[t]
	\centering
	\subfigure{\includegraphics[width=0.85\linewidth]{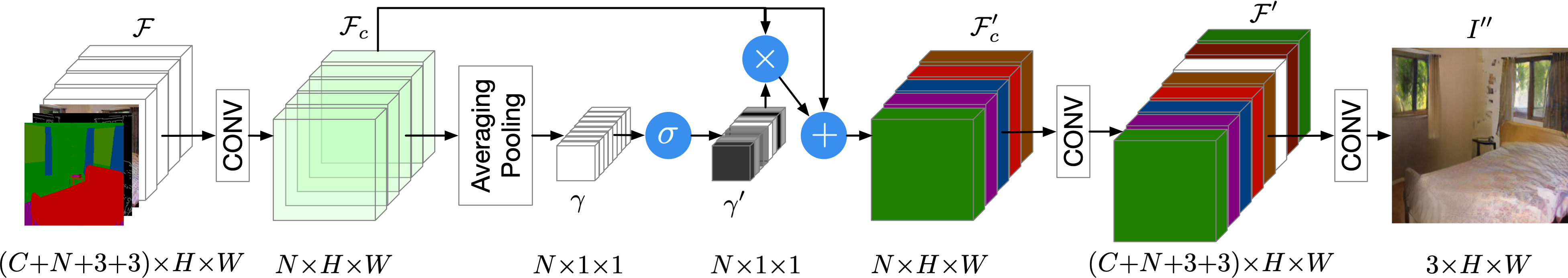}}
	\subfigure{\includegraphics[width=0.6\linewidth]{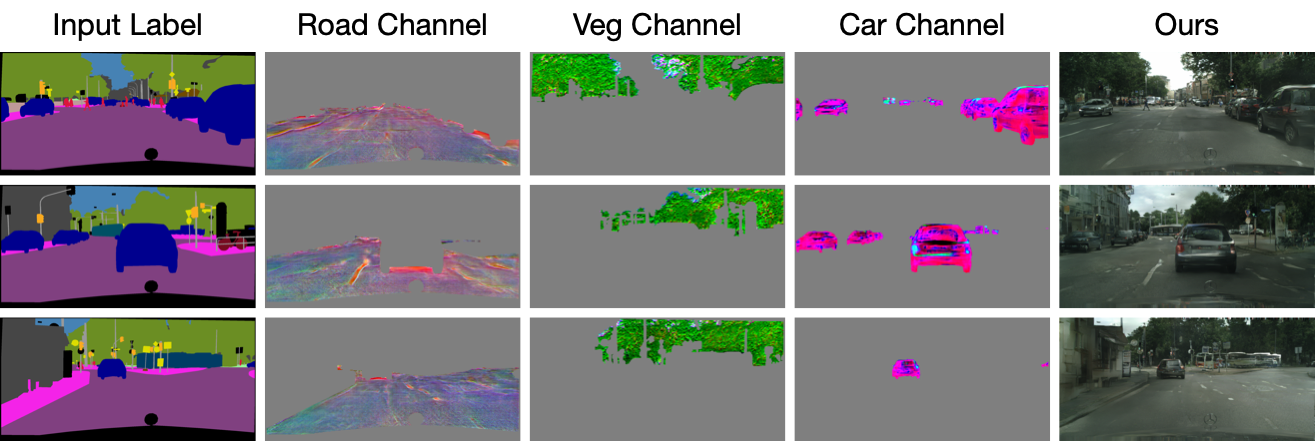}}
	\caption{\textbf{Top:} Overview of the proposed semantic preserving module $G_s$, which aims at capturing the semantic information and predicts scaling factors conditioned on the combined feature maps $\mathcal{F}$. These learned factors selectively highlight class-dependent feature maps, which are visualized in different colors. The symbols $\oplus$, $\otimes$, and ${\footnotesize \textcircled{$\sigma$}}$ denote element-wise addition, element-wise multiplication, and Sigmoid activation function, respectively. \textbf{Bottom:} Visualization of three different feature channels in $\mathcal{F}'$ on Cityscapes, i.e., road, car, and vegetation.}
	\label{fig:semantic_block}
		\vspace{-0.4cm}
\end{figure*}

\noindent \textbf{Edge Guided Image Generation.}
As discussed, the lack of detailed structure or geometry guidance makes it extremely difficult for the generator to produce realistic local structures and details.
To overcome this limitation, we propose to adopt the edge as guidance.
A novel edge generator $G_e$ is designed to directly generate the edge maps from the input semantic labels.
This also facilitates the shared encoder to learn more local structures of the targeted images.
Meanwhile, the image generator $G_i$ aims to generate photo-realistic images from the input labels.
In this way, the encoder is boosted to learn the appearance information of the targeted images.

Previous works \cite{park2019semantic,liu2019learning,qi2018semi,chen2017photographic,wang2018high} directly use deep networks to generate the target image, which is challenging since the network needs to simultaneously learn appearance and structure information from the input labels.
In contrast, the proposed method  learns structure and appearance separately via the proposed edge generator and image generator. 
Moreover, the explicit guidance from the ground truth edge maps can also facilitate the training of the encoder.
The framework of both edge and image generators is illustrated in Figure~\ref{fig:edge_block}.
Given the feature maps from the last convolutional layer of the encoder, i.e., $F {\in} \mathbb{R}^{C \times H \times W}$, 
where $H$ and $W$ are the width and height of the features, and $C$ is the number of channels, the edge generator produces edge features and edge maps which are further utilized to guide the image generator to generate the intermediate image $I'$.
The edge generator $G_e$ contains $n$ convolution layers and correspondingly produces $n$ intermediate feature maps $F_e {=} \{ F_e^j\}_{j=1}^n$.
After that, another convolution layer with Tanh non-linear activation is utilized to generate the edge map $I'_e{\in} \mathbb{R}^{3 \times H \times W}$.
Meanwhile, the feature maps $F$ is also fed into the image generator $G_i$ to generate $n$ intermediate feature maps $F_i{=}\{F_i^j\}_{j=1}^n$.
Then another convolution operation with Tanh non-linear activation is adopted to produce the intermediate image $I'_i{\in} \mathbb{R}^{3 \times H \times W}$.
In addition, the intermediate edge feature maps $F_e$ and the edge map $I'_e$ are utilized to guide the generation of the image feature maps $F_i$ and the intermediate image $I'$ via the Attention Guided Edge Transfer as detailed below.

\noindent \textbf{Attention Guided Edge Transfer.}
We further propose a novel attention guided edge transfer module $G_t$ to explicitly employ the edge structure information to refine the intermediate image representations.
The architecture of the proposed transfer module $G_t$ is illustrated in Figure~\ref{fig:edge_block}.
To transfer useful structure information from edge feature maps $F_e {=} \{ F_e^j\}_{j=1}^n$ to the image feature maps $F_i{=}\{F_i^j\}_{j=1}^n$, the edge feature maps are firstly processed by a Sigmoid activation function to generate the corresponding attention maps $F_a {=}{\rm Sigmoid}(F_e) {=} \{ F_a^j\}_{j=1}^n$.
The attention aims to provide structural information (which cannot be provided by the input label map) within each semantic class.
Then, we multiply the generated attention maps with the corresponding image feature maps to obtain the refined maps, which incorporate local structures and details.
Finally, the edge refined features are element-wisely summed with the original image features to produce the final edge refined  features, which are further fed to the next convolution layer as $F_i^j {=}{\rm Sigmoid}(F_e^j) {\times} F_i^j {+} F_i^j  (j  {=}  1, \cdots, n)$.
In this way, the image feature maps also contain the local structure information provided by the edge feature maps.
Similarly, to directly employ the structure information from the generated edge map $I_e^{'}$ for image generation, we adopt the attention guided edge transfer module to refine the generated image directly with edge information as
\begin{equation}
\begin{aligned}
I' = {\rm Sigmoid}(I'_e) \times I'_i +  I'_i,
\end{aligned}\label{eqn:image}
\end{equation}
where $I'_a{=}{\rm Sigmoid}(I'_e)$ is the generated attention map. We also visualize the results in Figure~\ref{fig:diff2}.

\begin{figure*} [!t]
	\centering
	\includegraphics[width=0.85\linewidth]{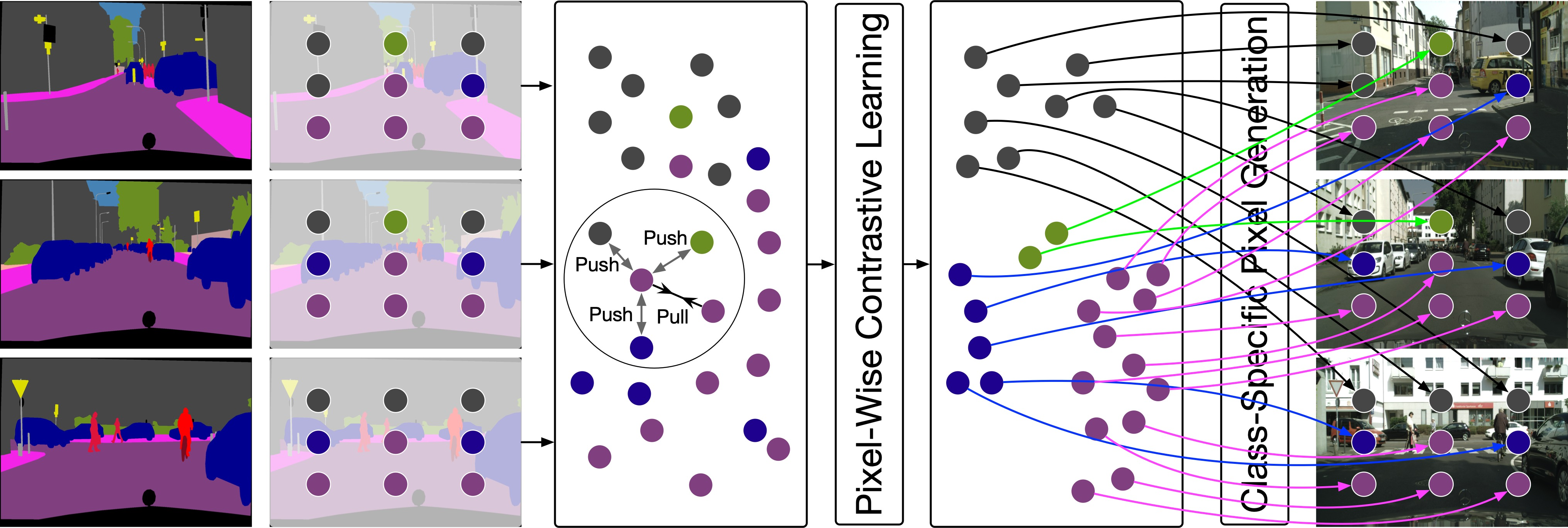}
	\caption{Current semantic image synthesis methods learn to map pixels to an embedding space but ignore intrinsic structures of labeled training data (i.e., inter-layout relations among pixels from the same class, marked with the same color). 
Our proposed approach, pixel-wise contrastive learning, fosters a new training strategy by explicitly addressing intra-class compactness and inter-class dispersion.
 By pulling pixels of the same class closer and pushing pixels from different classes apart, our method can create a better-structured embedding space, which leads to the same class generating more similar image content and improves the performance of semantic image synthesis.
	}
	\label{fig:method_contrastive}
		\vspace{-0.4cm}
\end{figure*}

\subsection{Semantic Preserving Image Enhancement}

\noindent \textbf{Semantic Preserving Module}. Due to the spatial resolution loss caused by  convolution, normalization, and down-sampling layers, existing models \cite{wang2018high,park2019semantic,qi2018semi,chen2017photographic} cannot fully preserve the semantic information of the input labels as illustrated in Figure~\ref{fig:city_seg}.
For instance, the small ``pole'' is missing, and the large ``fence'' is incomplete.
To tackle this problem, we propose a novel semantic preserving module, which aims to select class-dependent feature maps and further enhance it through the guidance of the original semantic layout. 
An overview of the proposed semantic preserving module $G_s$ is shown in Figure \ref{fig:semantic_block}(left).
Specifically, the input of the module denoted as $\mathcal{F}$,
is the concatenation of the input label $S$, the generated intermediate edge map $I'_e$ and image $I'$, and the deep feature $F$ produced from the shared encoder $E$.
Then, we apply a convolution operation on $\mathcal{F}$ to produce a new feature map $\mathcal{F}_c$ with the number of channels equal to the number of semantic categories, where each channel corresponds to a specific semantic category (a similar conclusion can be found in \cite{fu2019dual}).
Next, we apply the averaging pooling operation on $\mathcal{F}_c$ to obtain the global information of each class, followed by a  Sigmoid activation function to derive scaling factors $\gamma'$ as in $\gamma' {=} {\rm Sigmoid} ({\rm AvgPool}(\mathcal{F}_c))$, where each value represents the importance of the corresponding class. 
Then, the scaling factor $\gamma'$ is adopted to reweight the feature map $\mathcal{F}_c$ and highlight corresponding class-dependent feature maps.
The reweighted feature map is further added with the original feature $\mathcal{F}_c$ to compensate for information loss due to multiplication, and 
produces $\mathcal{F}'_c {=} \mathcal{F}_c  {\times} \gamma' {+} \mathcal{F}_c$, where $\mathcal{F}'_c {\in} \mathbb{R}^{N \times H \times W}$.

After that, we perform another convolution operation on $\mathcal{F}'_c$ to obtain the feature map $\mathcal{F}' {\in} \mathbb{R}^{(C+N+3+3) \times H \times W}$ to enhance the representative capability of the feature. In addition, $\mathcal{F}'$ has the same size as the original input one $\mathcal{F}$, which makes the module flexible and can be plugged into other existing architectures without modifications of other parts to refine the output.
In Figure~\ref{fig:semantic_block}(right), we visualize three channels in~$\mathcal{F}'$ on Cityscapes, i.e., road, car, and vegetation.
We can easily observe that each channel learns well the class-level deep
representations.

Finally, the feature map $\mathcal{F}'$ is fed into a convolution layer followed by a Tanh non-linear activation layer to obtain the final result $I''$.
Our semantic preserving module enhances the representational power of the model by adaptively recalibrating semantic class-dependent feature maps, and shares similar spirits with style transfer \cite{huang2017arbitrary}, and SENet \cite{hu2018squeeze} and EncNet \cite{zhang2018context}. 
One intuitive example of the utility of the module is for the generation of small object classes: these classes are easily missed in the generation results due to spatial resolution loss, while our scaling factor can put an emphasis on small objects and help preserve them.

\noindent \textbf{Similarity Loss.} 
Preserving semantic information from isolated pixels is very challenging for deep networks. To explicitly enforce the network to capture the relationship between semantic categories, a new similarity loss is introduced. This loss forces the network to consider both intra-class and inter-class pixels for each pixel in the label. Specifically, a state-of-the-art pretrained model (i.e., SegFormer \cite{xie2021segformer}) is used to transfer the generated image $I''$ back to a label $S'' {\in} \mathbb{R}^{N \times H \times W}$, where $N$ is the total number of semantic classes, and $H$ and $W$ represent the width and height of the image, respectively. A conventional method uses the cross entropy loss between $S''$ and $S$ to address this problem. 
However, such a loss only considers the isolated pixel while ignoring the semantic correlation with other pixels.

To address this limitation, we construct a similarity map from $S{\in} \mathbb{R}^{N \times H \times W}$. 
Firstly, we reshape $S$ to $\hat{S}{\in} \mathbb{R}^{N {\times} M}$, where $M {=} H {×}W$. Next, we perform a matrix multiplication to obtain a similarity map $A{=}\hat{S}\hat{S}^\top {\in} \mathbb{R}^{M{\times}M}$. This similarity map encodes which pixels belong to the same category, meaning that if the j-\textit{th} pixel and the i-\textit{th} pixel belong to the same category, then the value of the j-\textit{th} row and the i-\textit{th} column in $A$ is 1; otherwise, it is 0.
Similarly, we can obtain a similarity map $A''$ from the label $S''$.
Finally, we calculate the binary cross entropy loss between the two similarity maps $\{a_m {\in}A, m{\in} [1, M^2]\}$ and $\{a''_m{\in}A'', m{\in}[1, M^2]\}$ as
\begin{equation}
	\begin{aligned}
\mathcal{L}_{sim}(S, S'') = - \frac{1}{M^2} \sum_{m=1}^{M^2} (a_m \log a''_m + (1-a_m)\log (1-a''_m)).	\end{aligned}\label{eq:similarityloss}
\end{equation}
This loss explicitly captures intra-class and inter-class semantic correlation, leading to better generation results.

\begin{figure*} [!t]
	\centering
	\includegraphics[width=0.85\linewidth]{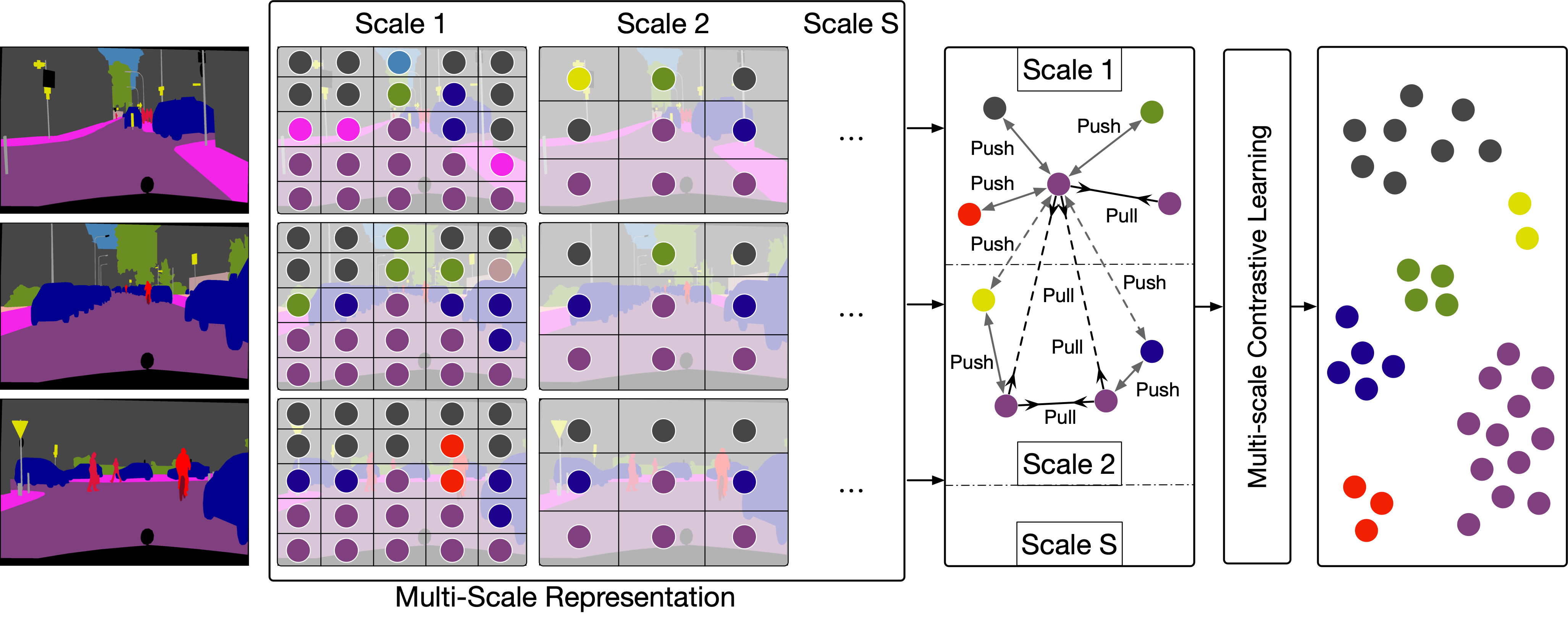}
	\caption{
 Our proposed multi-scale contrastive learning method is built upon the multi-scale features extracted from multiple input layouts. In addition, we introduce two novel multi-scale and cross-scale contrastive learning losses that are applied at multiple scale features and across scale features, within a shared feature space.
	}
	\label{fig:method_contrastive_multi}
		\vspace{-0.4cm}
\end{figure*}

\subsection{Contrastive Learning for Semantic Image Synthesis}

\noindent\textbf{Pixel-Wise Contrastive Learning.}
Existing semantic image synthesis models use deep networks to map labeled pixels to a non-linear embedding space. However, these models often only take into account the ``local'' context of pixel samples within an individual input semantic layout, and fail to consider the ``global'' context of the entire dataset, which includes the semantic relationships between pixels across different input layouts. This oversight raises an important question: what should the ideal semantic image synthesis embedding space look like? Ideally, such a space should not only enable accurate categorization of individual pixel embeddings, but also exhibit a well-structured organization that promotes intra-class similarity and inter-class difference. 
That is, pixels from the same class should generate more similar image content than those from different classes in the embedding space.
Previous approaches to representation learning propose that incorporating the inherent structure of training data can enhance feature discriminativeness. 
Hence, we conjecture that despite the impressive performance of existing algorithms, there is potential to create a more well-structured pixel embedding space by integrating both the local and global context.

The objective of unsupervised representation learning is to train an encoder that maps each training semantic layout $S$ to a feature vector $v {=} B(S)$, where $B$ represents the backbone encoder network. The resulting vector $v$ should be an accurate representation of $S$. To accomplish this task, contrastive learning approaches use a training method that distinguishes a positive from multiple negatives, based on the similarity principle between samples. The InfoNCE \cite{van2018representation,gutmann2010noise} loss function, a popular choice for contrastive learning, can be expressed as
\begin{equation}
		\mathcal{L}_S =   -\log
		\frac {\exp(v  \cdot  v_+ /  \tau  )}{
			\exp (v  \cdot  v_+  / \tau) +  \sum _ {v_- \in N_ {S}} {\exp(v\cdot v_- / \tau) }},
\end{equation}
where $v_+$ represents an embedding of a positive for $S$, and $N_S$ includes embeddings of negatives. The symbol ``·'' refers to the inner (dot) product, and $\tau {>}0$ is a temperature hyper-parameter. It is worth noting that the embeddings used in the loss function are normalized using the $L_2$ method.

One limitation of this training objective design is that it only penalizes pixel-wise predictions independently, without considering the cross-relationship between pixels. 
To overcome this limitation, we take inspiration from \cite{wang2021exploring,khosla2020supervised} and propose a contrastive learning method that operates at the pixel level and is intended to regularize the embedding space while also investigating the global structures present in the training data (see Figure \ref{fig:method_contrastive}).
Specifically, our contrastive loss computation uses training semantic layout pixels as data samples. For a given pixel $i$ with its ground-truth semantic label $c$, the positive samples consist of other pixels that belong to the same class $c$, while the negative samples include pixels belonging to other classes $C{\setminus}{c}$. As a result, the proposed pixel-wise contrastive learning loss is defined as follows
\begin{equation}
		\mathcal{L}_i =   \frac {1}{|P_ {i}|}   \sum_{i_+ \in P_i}  -\log
		\frac {\exp(i  \cdot  i_+ /  \tau  )}{
			\exp (i  \cdot  i_+  / \tau) +  \sum _ {i_- \in N_ {i}} {\exp(i\cdot i_- / \tau) }}.
\label{eq:contrastive}
\end{equation}
For each pixel $i$, we use $P_i$ and $N_i$ to represent the pixel embedding collections of positive and negative samples, respectively. Importantly, the positive and negative samples and the anchor $i$ are not required to come from the same layout. The goal of this pixel-wise contrastive learning approach is to create an embedding space in which same-class pixels are pulled closer together, and different-class pixels are pushed further apart. The result of this process is that pixels with the same class generate image contents that are more similar, which can lead to superior generation performance.

\noindent \textbf{Multi-Scale Contrastive Learning.}
In this part, we extend the pixel-level loss function $\mathcal{L}_i$ in Eq. \eqref{eq:contrastive} to an
arbitrary scale loss function $\mathcal{L}_i^s$ for calculating the contrastive learning loss, where $s$ means the $s$-\textit{th} scale feature representation, and we have a total of $\mathcal{S}$ different scales. 
This strategy regularizes the feature space of different scales by pulling features of the same class closer and pulling features of different classes apart, leading to a more well-structured feature space.

The overview framework of the proposed multi-scale contrastive learning is shown in Figure \ref{fig:method_contrastive_multi}. 
First, the input layouts go through the backbone encoder network $B$ to obtain multi-scale representation.
Next, we use a weighted sum at different scales to constraint the multi-scale features
\begin{equation}
\begin{aligned}
& \mathcal{L}_{i}^{ms}  =  \sum_{s=1}^\mathcal{S} w_s \mathcal{L}_i^s = w_1 \mathcal{L}_i^1 + \cdots + w_s \mathcal{L}_i^s + \cdots + w_\mathcal{S} \mathcal{L}_i^\mathcal{S} = \\
&  w_1 \frac {1}{|P_{i}^1|}   \sum_{i_+^1 \in P_i^1}  {-}\log \frac {\exp(i^1  \cdot  i_+^1 /  \tau  )}{\exp (i^1  \cdot  i_+^1  / \tau) +  \sum _ {i_-^1 \in N_{i}^1} {\exp(i^1\cdot i_-^1 / \tau) }} \\
& + \cdots + \\
&  w_s \frac {1}{|P_{i}^s|}   \sum_{i_+^s \in P_i^s}  {-}\log \frac {\exp(i^s  \cdot  i_+^s /  \tau  )}{\exp (i^s  \cdot  i_+^s  / \tau) +  \sum _ {i_-^s \in N_{i}^s} {\exp(i^s\cdot i_-^s / \tau) }} \\
   & + \cdots + \\
& w_\mathcal{S} \frac {1}{|P_{i}^\mathcal{S}|}   \sum_{i_+^\mathcal{S} \in P_i^\mathcal{S}} {-}\log \frac {\exp(i^\mathcal{S}  \cdot  i_+^\mathcal{S} /  \tau  )}{
\exp (i^\mathcal{S}  \cdot  i_+^\mathcal{S}  / \tau) +  \sum _ {i_-^\mathcal{S} \in N_{i}^\mathcal{S}} {\exp(i^\mathcal{S}\cdot i_-^\mathcal{S} / \tau) }}.
\label{eq:contrastive_multi}
\end{aligned}
\end{equation}
To identify the semantic classes in each pixel of different scale feature maps, we use the original input layout downsampled to the spatial dimensions.
We select the feature pairs with the same semantic label and at the same scale as positive pairs. On the contrary, we choose the feature pairs with different semantic labels and within the same scale as negative pairs.
Specifically, for each pixel $i^s$, we use $P_i^s$ and $N_i^s$ to represent the pixel embedding collections of positive and negative samples at the $s$-\textit{th} scale feature representation, respectively. Noth that the positive and negative samples and the anchor $i^s$ are from different layouts but the same scale feature embedding space. The weights $[w_1, \cdots, w_s, \cdots, w_\mathcal{S}]$ control the contribution of each scale to the overall loss.
Note that the first scale loss $\mathcal{L}_i^1$ is the same as the pixel-wise contrastive learning $\mathcal{L}_i$ in Eq. \eqref{eq:contrastive}.

As shown in Figure \ref{fig:method_contrastive_multi}, we also need to push same-class features from different scales closer together and pull different-class features apart.
For instance, if we have two scales $s_p$ and $s_q$, we hope features of the same class to be close on scales $s_p$ and $s_q$ ($s_p{\neq}s_q$), and features of different classes to be far apart on both scales $s_p$ and $s_q$.
That is, local features should describe parts of objects/regions of their global structure of the object and vice versa.
Thus the cross-scale contrastive learning loss can be formulated as
\begin{equation}
\begin{aligned}
&\mathcal{L}_{i}^{cs} =  \sum_{s_p=1}^{s_p=S} \sum_{s_q=1}^{s_q=S} w_{s_p, s_q} \mathcal{L}_i^{s_p, s_q} = \\
& w_{1, 2} \mathcal{L}_i^{1, 2} + \cdots  + w_{1, s} \mathcal{L}_i^{1, s} + \cdots + w_{1, \mathcal{S}} \mathcal{L}_i^{1, \mathcal{S}} + \cdots + w_{s, \mathcal{S}} \mathcal{L}_i^{s, \mathcal{S}}.
\end{aligned}
\label{eq:contrastive_cross}
\end{equation}
We downsample the original input layout into layouts of different scales on the spatial dimension so that we can obtain the semantic labels at each scale. We select the feature pairs with the same semantic label but at different scales as positive samples. In contrast, we select feature pairs with different semantic labels and at different scales as negative samples.
By doing so, we can achieve a bidirectional local-global consistency for learning the encoder network.
The weights $[w_{1,2}, \cdots, w_{1,s}, \cdots, w_{1,\mathcal{S}}, \cdots, w_{s, \mathcal{S}}]$ control the contribution of each scale to the overall loss.

Eq. \eqref{eq:contrastive_multi} and \eqref{eq:contrastive_cross} can be added together to obtain our complete contrastive learning loss.

\noindent \textbf{Class-Specific Pixel Generation.}
To overcome the challenges posed by training data imbalance between different classes and size discrepancies between different semantic objects, we introduce a new approach that is specifically designed to generate small object classes and fine details. Our proposed method is a class-specific pixel generation approach that focuses on generating image content for each semantic class. Doing so can avoid the interference from large object classes during joint optimization, and each subgeneration branch can concentrate on a specific class generation, resulting in similar generation quality for different classes and yielding richer local image details.

An overview of the class-specific pixel generation method is provided in Figure~\ref{fig:method_contrastive}. 
After the proposed pixel-wise contrastive learning, we obtain a class-specific feature map for each pixel. 
Then, the feature map is fed into a decoder for the corresponding semantic class, which generates an output image $\hat{I}_{i}$. 
Since we have the proposed contrastive learning loss, we can use the parameter-shared decoder to generate all classes.
To better learn each class, we also utilize a pixel-wise $L_1$ reconstruction loss, which can be expressed as $\mathcal{L}_{L_1} {=} \sum_{i=1}^{N} \mathbb{E}_{I_i, \hat{I}_i} \lbrack \vert\vert I_i {-} \hat{I}_i \vert\vert_1 \rbrack.$
The final output $I_g$ from the pixel generation network can be obtained by performing an element-wise addition of all the class-specific outputs:
$I_g {=} I_{g_1} \oplus I_{g_2} \oplus \cdots \oplus I_{g_N}.$

\subsection{Model Training}
\noindent \textbf{Multi-Modality Discriminator.}
To facilitate the training of the proposed method for high-quality edge and image generation, a novel multi-modality discriminator is developed to simultaneously distinguish outputs from two modality spaces, i.e., edge and image. 
Since the edges and RGB images share the same structure, they can be learned using the multi-modality discriminator. In the preliminary experiment, we also tried to use two discriminators (i.e., an edge discriminator and an image discriminator), but no performance improvement was observed while increasing the model complexity. Thus, we use the proposed multi-modality discriminator.
The framework of the multi-modality discriminator is shown in Figure~\ref{fig:method}, which is capable of discriminating both real/fake images and edges. 
To discriminate real/fake edges, the discriminator loss considering the semantic label $S$ and the generated edge $I'_e$ (or the real edge $I_e$) is as
\begin{equation}
\begin{aligned}
\mathcal{L}_{\mathrm{CGAN}}(G_e, D) & = 
\mathbb{E}_{S, I_e} \left[ \log D(S, I_e) \right] \\
& +  \mathbb{E}_{S, I'_e} \left[\log (1 - D(S, I'_e)) \right],
\end{aligned}
\label{eqn:discriminator1}
\end{equation}
which guides the model to distinguish real edges from fake generated edges.
Further, to discriminate real/fake images, the discriminator loss regarding  semantic label $S$ and the generated images $I'$, $I''$ (or the real image $I$) is as Eq.~\eqref{eqn:discriminator2}, which guides the model to discriminate real/fake images,
\begin{equation}
	\begin{aligned}
\mathcal{L}_{\mathrm{CGAN}}(G_i, G_s, D)  & = (\lambda + 1) \mathbb{E}_{S, I} \left[ \log D(S, I) \right]  \\
& +  \mathbb{E}_{S, I'} \left[\log (1 - D(S, I')) \right] \\
& +  \lambda \mathbb{E}_{S, I''} \left[\log (1 - D(S, I'')) \right],
\end{aligned}
\label{eqn:discriminator2}
\end{equation}
where $\lambda$ controls the losses of the two generated images. 
The inclusion of $I'$ and $I''$ is a cascaded coarse-to-fine generation strategy \cite{tang2019multi}, i.e., $I'$ is the coarse result, while $I''$ is the refined result. 
The intuition is that $I''$ will be better generated based on $I'$, so we provide $I'$ to the discriminator to ensure that $I'$ is also realistic. 

\begin{figure*} [!t] \small
	\centering
	\includegraphics[width=1\linewidth]{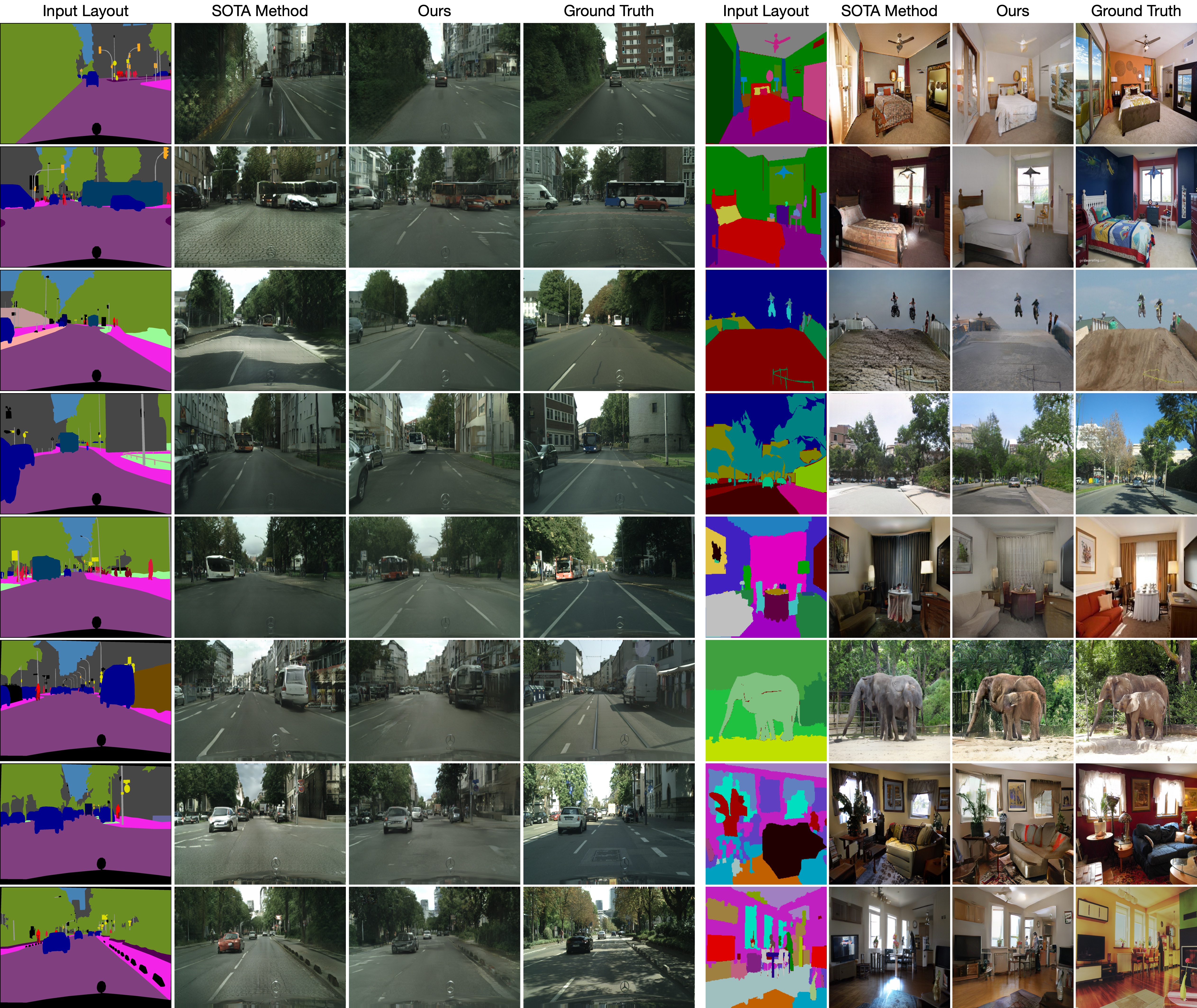}
	\caption{Existing state-of-the-art method (i.e., OASIS) vs. our proposed ECGAN on three datasets. 
	Cityscapes: left; ADE20K: top right four; COCO-Stuff: bottom right four.}
	\label{fig:sota}
		\vspace{-0.2cm}
\end{figure*}

\noindent \textbf{Optimization Objective.}
Equipped with the multi-modality discriminator, we elaborate on the training objective for the proposed method as follows.
Five different losses, i.e., the multi-modality adversarial loss, the similarity loss, the contrastive learning loss, the discriminator feature matching loss $\mathcal{L}_{f}$, and the perceptual loss $\mathcal{L}_{p}$ are used to optimize the proposed ECGAN,
\begin{equation}
\begin{aligned}
\min_{G} \max_{D} \mathcal{L} & = \lambda_{c} \underbrace{(\mathcal{L}_{\mathrm{CGAN}}(G_e, D) 
+ \mathcal{L}_{\mathrm{CGAN}}(G_i, G_s, D))}_{\text{Multi-Modality Adversarial Loss}} \\
&  + \lambda_{s} \underbrace{\mathcal{L}_{sim}(S, S')  + \mathcal{L}_{sim}(S, S'')}_{\text{Similarity Loss}} \\
& +  \lambda_{l} \underbrace{\mathcal{L}_{i}^{ms} + \mathcal{L}_{i}^{cs} + \mathcal{L}_{L_1}}_{\text{Contrastive Learning Loss}}  \\
& + \lambda_{f}\underbrace{(\mathcal{L}_{f}(I_e, I'_e) {+} \mathcal{L}_{f}(I, I') {+} \lambda \mathcal{L}_{f}(I, I''))}_{\text{Discriminator  Feature Matching Loss}} \\
& + \lambda_{p} \underbrace{(\mathcal{L}_{p}(I_e, I'_e) {+} \mathcal{L}_{p}(I, I') {+} \lambda \mathcal{L}_{p}(I, I''))}_{\text{Perceptual Loss}},
\label{eq:loss} 
\end{aligned}
\end{equation}
where $\lambda_{c}$, $\lambda_{s}$, $\lambda_{l}$, $\lambda_{f}$, and $\lambda_{p}$ are the parameters of the corresponding loss that contributes to the total loss $\mathcal{L}$;
where $\mathcal{L}_{f}$ matches the discriminator intermediate features between the generated images/edges and the real images/edges; where $\mathcal{L}_{p}$ matches the VGG extracted features between the generated images/edges and the real images/edges.
By maximizing the discriminator loss, the generator is promoted to simultaneously  generate reasonable edge maps that can capture the local-aware structure information and generate realistic images semantically aligned with the input labels.

\subsection{Implementation Details}
\label{sm:Implementation}

For both the image generator $G_i$ and edge generator $G_e$, the kernel size and padding size of convolution layers are all $3 {\times} 3$ and 1 for preserving the feature map size.
We 	set $n{=}3$ for  generators $G_i$, $G_s$, and $G_t$.
The channel size of feature $F$ is set to $C{=}64$. 
For the semantic preserving module $G_s$, we adopt an adaptive average pooling operation.
Spectral normalization \cite{miyato2018spectral} is applied to all the layers in both the generator and discriminator.
Our method incorporates the use of the Canny edge detector \cite{canny1986computational} for the purpose of deriving edge maps essential to our training process. In the subsequent testing phase, our approach necessitates no supplemental data, maintaining the fairness of comparisons with other existing methods.

\begin{table*}[!t] \small
	\centering
	\caption{User study on Cityscapes, ADE20K, and COCO-Stuff. The numbers indicate the percentage of users who favor the results of the proposed ECGAN over the competing methods.}
%		\resizebox{1\linewidth}{!}{% 
	\begin{tabular}{lccc} \toprule
		AMT $\uparrow$                               & Cityscapes & ADE20K  &  COCO-Stuff \\ \midrule
		Our ECGAN vs. CRN~\cite{chen2017photographic}     & 88.8 {$\pm$ 3.4}      & 94.8 {$\pm$ 2.7} & 95.3 {$\pm$ 2.1}\\
		Our ECGAN vs. Pix2pixHD~\cite{wang2018high}        & 87.2 {$\pm$ 2.9}       & 93.6 {$\pm$ 3.1}  &  93.9 {$\pm$ 2.4} \\ 
		Our ECGAN vs. SIMS~\cite{qi2018semi}                      & 85.3 {$\pm$ 3.8}       & -     & - \\
		Our ECGAN vs. GauGAN~\cite{park2019semantic}    & 84.7 {$\pm$ 4.3}       & 88.4 {$\pm$ 3.7}  &  90.8 {$\pm$ 2.5} \\ 
		Our ECGAN vs. DAGAN~\cite{tang2020dual}             & 81.8 {$\pm$ 3.9}       & 86.2 {$\pm$ 3.6}  & -\\
		Our ECGAN vs. CC-FPSE~\cite{liu2019learning}         & 79.5 {$\pm$ 4.2}       & 85.1 {$\pm$ 3.9}  &  86.7 {$\pm$ 2.8} \\  
		Our ECGAN vs. LGGAN \cite{tang2020local}              & 78.4 {$\pm$ 4.7}       & 82.7 {$\pm$ 4.5} & - \\
		Our ECGAN vs. OASIS \cite{sushko2020you}             & 76.7 {$\pm$ 4.8}        & 80.6 {$\pm$ 4.5}  & 82.5 {$\pm$ 3.1}  \\	\bottomrule
	\end{tabular}
	\label{tab:atm1}
		\vspace{-0.2cm}
\end{table*}

\begin{table*}[!t] \small
	\centering
	\caption{User study on Cityscapes, ADE20K, and COCO-Stuff. The numbers indicate the percentage of users who favor the results of the proposed ECGAN++ over the proposed ECGAN.}
%		\resizebox{1\linewidth}{!}{% 
	\begin{tabular}{lccc} \toprule
		AMT $\uparrow$                               & Cityscapes & ADE20K  &  COCO-Stuff \\ \midrule
		Our ECGAN++ vs. Our ECGAN \cite{tang2023edge}           & 64.3 {$\pm$ 3.2}        & 67.5 {$\pm$ 3.8}  & 70.4 {$\pm$ 2.6}  \\	\bottomrule
	\end{tabular}
	\label{tab:atm2}
		\vspace{-0.2cm}
\end{table*}

\begin{table*}[!t] \small
	\centering
	\caption{Quantitative comparison of different methods on Cityscapes, ADE20K, and COCO-Stuff.
	}
	\resizebox{1\linewidth}{!}{% 
	\begin{tabular}{rlllllllll} \toprule
		\multirow{2}{*}{Method}  & \multicolumn{3}{c}{Cityscapes} & \multicolumn{3}{c}{ADE20K} & \multicolumn{3}{c}{COCO-Stuff} \\ \cmidrule(lr){2-4} \cmidrule(lr){5-7} \cmidrule(lr){8-10} 
		& mIoU $\uparrow$    & Acc $\uparrow$  & FID  $\downarrow$ & mIoU $\uparrow$    & Acc $\uparrow$  & FID  $\downarrow$  & mIoU $\uparrow$    & Acc $\uparrow$  & FID  $\downarrow$ \\ \midrule
		CRN~\cite{chen2017photographic}  & 52.4  & 77.1 & 104.7  & 22.4 & 68.8 & 73.3 & 23.7 & 40.4 & 70.4\\
		SIMS~\cite{qi2018semi}           & 47.2  & 75.5 & 49.7 & - & - & - & - & - & - \\
		Pix2pixHD~\cite{wang2018high}    & 58.3  & 81.4 & 95.0  & 20.3 & 69.2 & 81.8  & 14.6 & 45.8 & 111.5  \\ 
  		GauGAN~\cite{park2019semantic}   & 62.3  & 81.9 & 71.8  & 38.5 & 79.9 & 33.9 & 37.4 & 67.9 & 22.6\\
         DPGAN \cite{tang2021layout} & 65.2 & 82.6 & 53.0 & 39.2 & 80.4 & 31.7 & - & - & - \\
		DAGAN \cite{tang2020dual} & 66.1 & 82.6 & 60.3 &   40.5 &  81.6 & 31.9 & - & - & -\\
            SelectionGAN \cite{tang2019multi} & 83.8 & 82.4 & 65.2 & 40.1 & 81.2 & 33.1 & - & - & -\\
            SelectionGAN++ \cite{tang2022multi} & 64.5 & 82.7 & 63.4 & 41.7 & 81.5 & 32.2 & - & - & - \\
		LGGAN \cite{tang2020local} & 68.4 & 83.0 & 57.7 & 41.6 & 81.8 & 31.6 & - & - & -\\
            LGGAN++ \cite{tang2022local}  & 67.7 & 82.9 & 48.1 & 41.4 & 81.5 & 30.5 & - & - & - \\
		CC-FPSE~\cite{liu2019learning}  & 65.5 & 82.3 & 54.3 & 43.7 & 82.9 & 31.7 & 41.6 & 70.7 & 19.2 \\
            \hao{SCG \cite{wang2021image}} & \hao{66.9} & \hao{82.5} & \hao{49.5} & \hao{45.2} & \hao{83.8} & \hao{29.3} & \hao{42.0} & \hao{72.0} & \hao{18.1}\\
		OASIS \cite{sushko2020you} &   69.3 & - & 47.7 & 48.8 & - & 28.3 & 44.1 & - & 17.0  \\
		\hao{RESAIL \cite{shi2022retrieval}} & \hao{69.7} & \hao{83.2}
		& \hao{45.5} & \hao{49.3} & \hao{84.8} &\hao{30.2} & \hao{44.7} & \hao{73.1} & \hao{18.3} \\
		\hao{SAFM \cite{lv2022semantic}} &\hao{70.4} & \hao{83.1} &\hao{49.5} &\hao{50.1}&\textbf{\hao{86.6}}&\hao{32.8}& \hao{43.3} & \textbf{\hao{73.4}} & \hao{24.6} \\
            \hao{PITI \cite{wang2022pretraining}} & \hao{-} & \hao{-} & \hao{-} & \hao{-} & \hao{-} & \hao{-} & \hao{-} & \hao{-} & \hao{19.36}\\
            \hao{T2I-Adapter \cite{mou2023t2i}} & \hao{-} & \hao{-} & \hao{-} & \hao{-} & \hao{-} & \hao{-} & \hao{-} & \hao{-} & \hao{16.78}\\
            \hao{SDM \cite{wang2022semantic}} & \hao{-} & \hao{-} & \hao{\textbf{42.1}} & \hao{-} & \hao{-} & \hao{27.5} & \hao{-} & \hao{-} & \hao{15.9}\\
		ECGAN (Ours)                        & 72.2    & 83.1 & 44.5 & 50.6 & 83.1 & 25.8 & 46.3 & 70.5 & 15.7 \\
		ECGAN++ (Ours) & \textbf{73.3} (+1.1) & \textbf{83.9} (+0.8) & 42.2 (-2.3) & \textbf{52.7} (+2.1) & 85.9 (+2.8) & \textbf{24.7} (-1.1) & \textbf{47.9} (+1.6) & 72.3 (+1.8) & \textbf{14.9} (-0.8) \\
		\bottomrule
	\end{tabular}}
	\label{tab:sota}
	\vspace{-0.4cm}
\end{table*}

In our computation of the contrastive learning loss, we observe a direct correlation between the number of layouts used and the resultant performance, i.e., more layouts lead to enhanced performance. However, a plateau is reached when the count exceeds 8 layouts; additional layouts contribute only marginal improvements to performance, while significantly slowing down the overall training process. Thus, with the objective of striking a balance between performance efficiency and computational time, we elect to use 8 layouts as input for the calculation of contrastive learning loss.
We use features from four scales in Eq. \eqref{eq:contrastive_multi}, with feature map output strides of 1, 4, 8, and 16, to calculate the multi-scale contrastive learning loss. 
Meanwhile, we also downsample the input layout by 4, 8, and 16 times to obtain the label of the corresponding scale for calculating the multi-scale contrastive learning loss.
The weights $w_s$ in Eq. \eqref{eq:contrastive_multi} are set to  1, 0.7, 0.4, and 0.1 in a decreasing way for feature maps of strides 1, 4, 8, and 16, respectively.
Moreover, in order to balance the performance and efficiency, we adopt two cross-scale contrastive learning in Eq. \eqref{eq:contrastive_cross}, i.e., (s4, s8) and (s4, s16).
We set both weights in Eq. \eqref{eq:contrastive_cross} to 0.1.

Also, we follow the training procedures of GANs \cite{goodfellow2014generative} and alternatively train the generator $G$ and discriminator $D$, i.e., one gradient descent step on the discriminator and generator alternately. 
We use the Adam solver \cite{kingma2014adam} and set $\beta_1{=}0$, $\beta_2{=}0.999$.
$\lambda_{c}$,  $\lambda_{s}$, $\lambda_{l}$, $\lambda_{f}$, and $\lambda_{p}$ in Eq.~\eqref{eq:loss} is set to 1, 1, 1, 10, and 10, respectively.
All $\lambda$ in both Eq.~\eqref{eqn:discriminator2} and \eqref{eq:loss} are set to 2.
We conduct experiments on an NVIDIA DGX1 with 8 V100 GPUs. 

\section{Experiments}

\begin{figure*} [!t] \small
	\centering
	\includegraphics[width=0.9\linewidth]{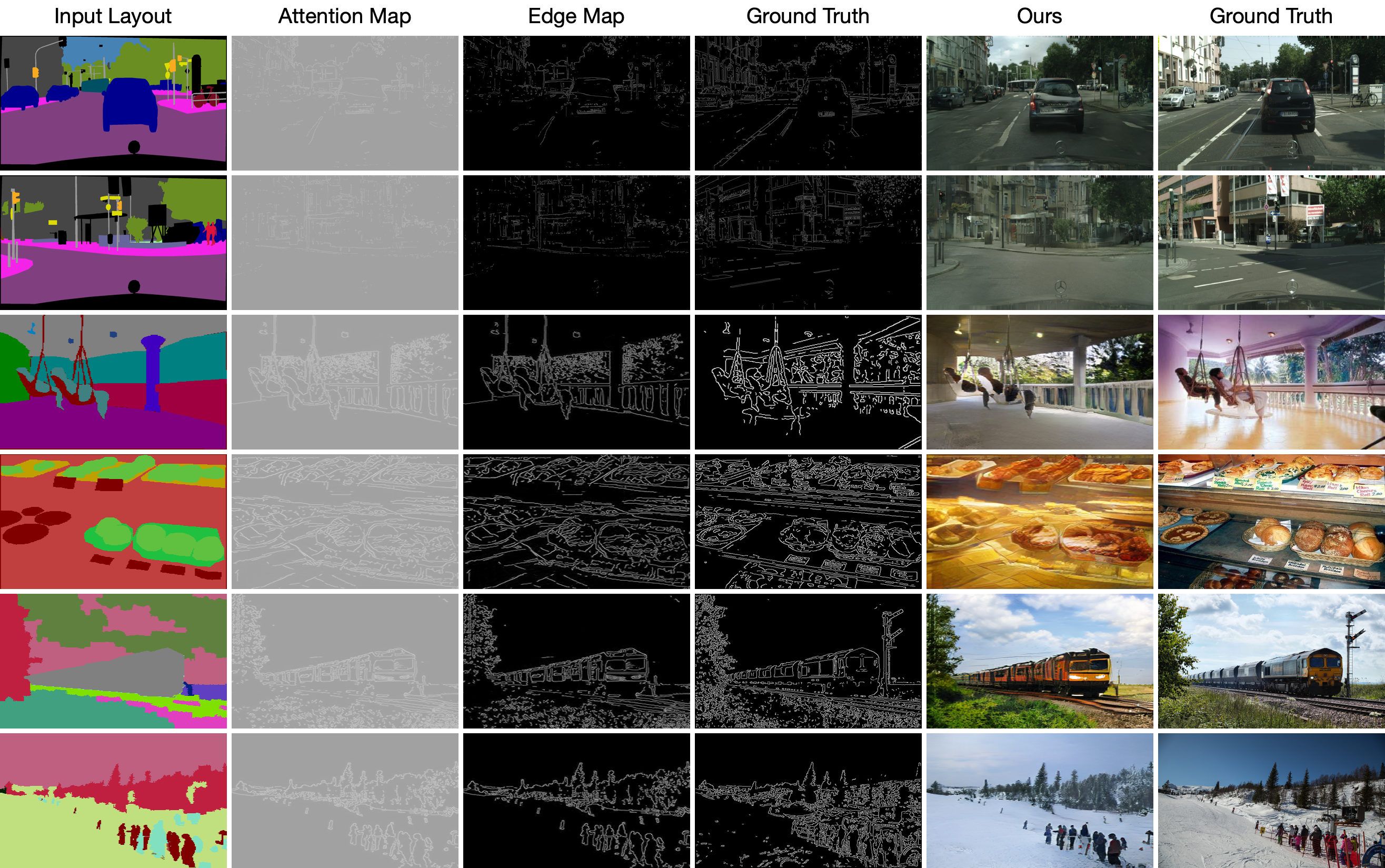}
	\caption{Edge and attention maps generated by the proposed method.
	}
	\label{fig:ade_gaugan}
	\vspace{-0.4cm}
\end{figure*}

\subsection{Experimental Setups} 
\noindent\textbf{Datasets.}
We follow GauGAN \cite{park2019semantic} and conduct experiments on three datasets, i.e., Cityscapes \cite{cordts2016cityscapes}, ADE20K \cite{zhou2017scene}, and COCO-Stuff \cite{caesar2018coco}.
For more detail about these datasets, please refer to GauGAN \cite{park2019semantic}.

\noindent\textbf{Evaluation Metrics.}
We employ the mean Intersection-over-Union (mIoU), Pixel Accuracy (Acc), and Fr\'echet Inception Distance (FID)~\cite{heusel2017gans} as the evaluation metrics.
For more detail about these evaluation metrics, please refer to GauGAN \cite{park2019semantic}.

\subsection{Experimental Results} 

\noindent \textbf{Qualitative Comparisons.}
We adopt GauGAN as the encoder $E$ to validate the effectiveness of the proposed method.
Visual comparison results on all three datasets with the state-of-the-art method (i.e., OASIS \cite{sushko2020you}) are shown in Figure~\ref{fig:sota}. 
\hao{
We can see that ECGAN and ECGAN++ achieve visually better results with fewer visual artifacts than the existing state-of-the-art method.
Examining Figure \ref{fig:sota}, it is evident that the SOTA method produces numerous visual artifacts across varied categories like vegetation, cars, buses, roads, buildings, fences, beds, cabinets, curtains, elephants, etc. In contrast, our approach generates significantly more realistic content, as can be observed on both sides of the figure. Moreover, the proposed methods generate more local structures and details than the SOTA method.}

\noindent\textbf{User Study.} 
We follow the same evaluation protocol as GauGAN and conduct a user study.
Specifically, we provide the participants with an input layout and two generated images from different models and ask them to choose the generated image that looks more like a corresponding image of the layout. 
The users are given unlimited time to make the decision. 
For each comparison, we randomly generate 400 questions for each dataset, and each question is answered by 10 different participants. For other methods, we use the public code and pretrained models provided by the authors to generate images.
As shown in Table \ref{tab:atm1}, users favor our synthesized results on all three datasets compared with other competing methods, further validating that the generated images by ECGAN are more natural.
Moreover, we can see in Table \ref{tab:atm2} that users favor our synthesized results by the proposed ECGAN++ compared with the proposed ECGAN, validating the effectiveness of the proposed multi-scale
contrastive learning method.

\noindent \textbf{Quantitative Comparisons.}
Although the user study is more suitable for evaluating the quality of the generated images, we also follow previous works and use mIoU, Acc, and FID for quantitative evaluation.
The results of the three datasets are shown in Table~\ref{tab:sota}. 
The proposed ECGAN and ECGAN++ outperform other leading methods by a large margin on all three datasets, validating the effectiveness of the proposed methods.

\noindent\hao{\textbf{Memory Usage.} The proposed method is memory-efficient compared to those methods which model the generation of different image classes individuals such as LGGAN \cite{tang2020local}. Thus, we compare the memory usage during training/testing when the batch size is set to 1. The memory (GB) of LGGAN on CityScapes (30 categories),  ADE20K (150 categories), and COCO-Stuff (182 categories) datasets are about 17.8, 23.9, and 28.1, respectively. The memory (GB) of our proposed method on the Cityscapes, ADE20K, and COCO-Stuff datasets is about 6.3, 5.6, and 5.9 respectively. It is clear that LGGAN's memory requirement significantly escalates as category numbers increase, whereas our method maintains comparable memory demands. This advantage becomes even more prominent when using larger batch sizes, implying we can train/test the model with larger batches on the same GPU devices.}

\noindent \textbf{Visualization of Edge and Attention Maps.}
We also visualize the generated edge and attention maps in Figure~\ref{fig:ade_gaugan}.
We observe that the proposed method can generate reasonable edge maps according to the input labels. Thus the generated edge maps can provide more local structure information for generating more photo-realistic images.

\begin{figure*} [!t] \small
	\centering
	\includegraphics[width=0.9\linewidth]{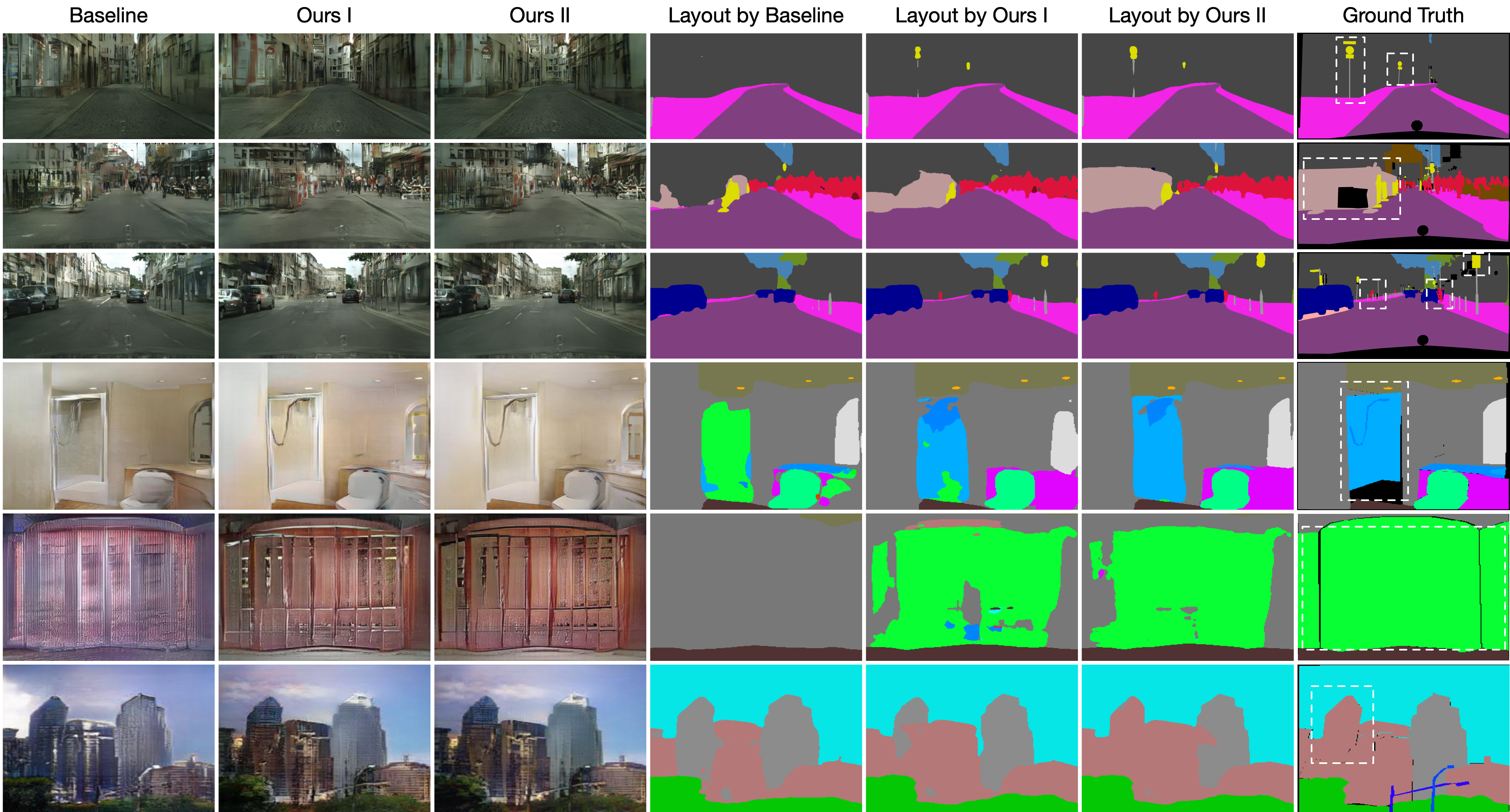}
	\caption{Segmentation layouts generated by the baseline  and our proposed method. ``Ours I'' and ``Ours II'' stand for $I'$ and $I''$, respectively.
	}
	\label{fig:city_seg}
		\vspace{-0.4cm}
\end{figure*}

\noindent \textbf{Visualization of Segmentation Maps.}
We follow GauGAN and apply pre-trained segmentation  networks \cite{yu2017dilated,xiao2018unified} on the generated images to produce segmentation maps.
Results compared with the baseline method are shown in Figure~\ref{fig:city_seg}.
We observe that the proposed method consistently generates better semantic labels than the baseline on both datasets.

\begin{table}[!t] \small
	\centering
 	\caption{Multi-modal synthesis evaluation on ADE20K.}
	\begin{tabular}{rccc} \toprule
		{Method} & {Multi-Modal} &  {LPIPS $\uparrow$}  \\ \midrule
		{GauGAN+ \cite{park2019semantic}} & {Encoder} & {0.16} \\
		{GauGAN+ \cite{park2019semantic}} & {3D Noise} & {0.50} \\
		{OASIS \cite{sushko2020you}} & 
		{3D Noise} & {0.35} \\
		ECGAN (Ours) & Encoder & 0.18 \\ 
		ECGAN (Ours) & 3D Noise & 0.52 \\
        ECGAN++ (Ours) & Encoder & 0.22\\ 
        ECGAN++ (Ours) & 3D Noise & \textbf{0.54}\\ \bottomrule
	\end{tabular}
 \vspace{-0.4cm}
	\label{tab:multimodal}
\end{table}

\begin{figure*}[!t] \small
	\centering
	\includegraphics[width=0.9\linewidth]{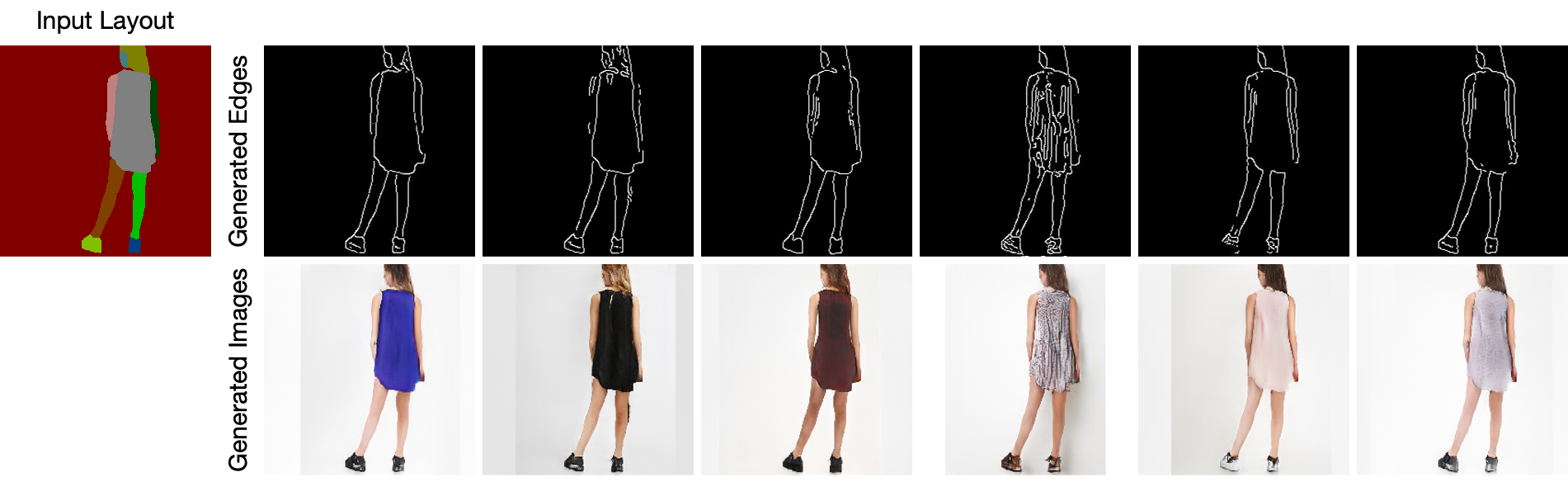}
	\caption{Results generated by the proposed method for multi-modal image and edge synthesis.}
	\label{fig:diff}
		\vspace{-0.2cm}
\end{figure*}

\noindent \textbf{Multi-Modal Image and Edge Synthesis.}
We follow GauGAN \cite{park2019semantic} and apply a style encoder and a KL-Divergence loss with a loss weight of 0.05 to enable multi-modal image and edge synthesis.
As shown in Figure~\ref{fig:diff}, our model generates different edges and images from the same input layout, which we believe will benefit other tasks, e.g., image restoration \cite{shi2022rcrn}, and image/video super-resolution \cite{wu2022compiler,cao2022towards}.
Moreover, we follow OASIS \cite{sushko2020you} and use LPIPS \cite{zhang2018unreasonable} to evaluate the variation in the multi-model image synthesis on the ADE20K dataset.
Following in OASIS, we generate 20 images and compute the mean pairwise scores, and then average over all label maps. 
The higher the LPIPS scores, the more diverse the generated images are. 
We follow OASIS and GauGAN, and employ two settings (i.e., encoder and 3D noise) to evaluate multi-modal image synthesis.
Table \ref{tab:multimodal} shows that the proposed ECGAN and ECGAN++ achieve better results than OASIS and GauGAN in both settings.
Note that existing methods (e.g., OASIS \cite{sushko2020you} and GauGAN \cite{park2019semantic}) can only achieve multi-modal image synthesis.

\begin{figure*}[!t] \small
	\centering
	\includegraphics[width=0.9\linewidth]{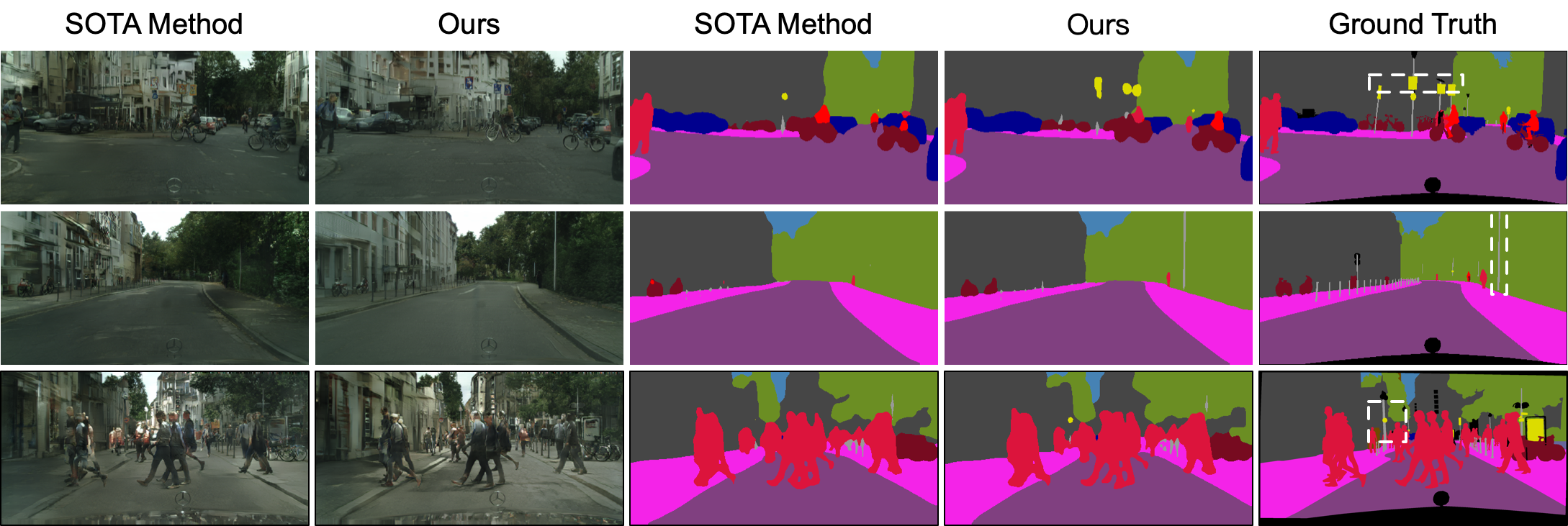}
	\caption{Visualization of small object generation on Cityscapes.}
	\label{fig:small}
         \vspace{-0.2cm}
\end{figure*}

\begin{table*}[!t]\small
	\centering
 	\caption{mIoU of small objects on Cityscapes.}
	% \resizebox{1\linewidth}{!}{% 
	\begin{tabular}{lccccccc}	\toprule
		{mIoU $\uparrow$}                               & {Pole} & {Light}  & {Sign} & {Rider}  & {Mbike} & {Bike}  & {Overall} \\ \midrule
		{OASIS} \cite{sushko2020you}  & {23.4} & {32.6} & {14.9} & {27.3} & {31.2} & {26.6} & {26.0} \\
		ECGAN (Ours) & 26.2 & 36.7 & 17.4 & 30.2 & 33.5 & 28.7 & 28.8 \\ 
        ECGAN++ (Ours) & \textbf{26.7} & \textbf{37.0} & \textbf{17.9} & \textbf{30.8} & \textbf{34.2} & \textbf{29.5} & \textbf{29.4} \\\bottomrule
	\end{tabular}
 \vspace{-0.2cm}
	\label{tab:small}
\end{table*}

\begin{table*}[!t] \small
	\centering
 	\caption{Ablation study of the proposed method on Cityscapes, ADE20K, and COCO-Stuff.}
	\resizebox{1\linewidth}{!}{% 
		\begin{tabular}{clccccccccc} \toprule
           \multirow{2}{*}{\#} & \multirow{2}{*}{Setting}  & \multicolumn{3}{c}{Cityscapes} & \multicolumn{3}{c}{ADE20K} & \multicolumn{3}{c}{COCO-Stuff} \\ \cmidrule(lr){3-5} \cmidrule(lr){6-8} \cmidrule(lr){9-11} 
		  &	&  mIoU $\uparrow$ & Acc $\uparrow$ & FID $\downarrow$  & {mIoU $\uparrow$} & {Acc $\uparrow$} & {FID $\downarrow$}  &  {mIoU $\uparrow$} & {Acc $\uparrow$} & {FID $\downarrow$} \\ \midrule	
		B1& $E$+$G_i$                    &  58.6      & 81.4      & 65.7 & 36.9 & 78.5 & 38.2 & 36.8 & 65.1 & 24.5 \\
		B2& $E$+$G_i$+$G_e$              &  60.2    & 81.7     & 61.0  & {38.7} & {79.2} & {36.3} & {37.5} & {66.3} & {22.9}\\
		B3& $E$+$G_i$+$G_e$+$G_t$        &  61.5     & 82.0     & 59.0  & {40.6} & {80.3} & {34.6} & {39.1} & {67.0} & {21.7}\\
		B4& $E$+$G_i$+$G_e$+$G_t$+$G_s$  &  64.5       & 82.5     & 57.1 & {42.0} & {82.0} & {32.4} & {41.4} & {68.2} & {19.8} \\  
		B5&	$E$+$G_i$+$G_e$+$G_t$+$G_s$+$G_l$ & 66.8 & 82.7 & 52.2 & {45.8} & {82.4} & {29.9} & {43.7} & {69.1} & {17.6} \\ 
		B6&	$E$+$G_i$+$G_e$+$G_t$+$G_s$+$G_l$+$G_c$ & 72.2 & 83.1 & 44.5 & {50.6} & {83.1} & {25.8} & {46.3} & {70.5} & {15.7} \\ 
  B7& $E$+$G_i$+$G_e$+$G_t$+$G_s$+$G_l$+$G_c$+Eq. \eqref{eq:contrastive_multi} & 72.8 & 83.5 & 43.7 & 51.6 & 84.3 & 25.3 & 47.1 & 71.4 & 15.4 \\
		B8& $E$+$G_i$+$G_e$+$G_t$+$G_s$+$G_l$+$G_c$+Eq. \eqref{eq:contrastive_multi}+Eq. \eqref{eq:contrastive_cross}  & \textbf{73.3} & \textbf{83.9} & \textbf{42.2} & \textbf{52.7} & \textbf{85.9} & \textbf{24.7} & \textbf{47.9} & \textbf{72.3} & \textbf{14.9} \\ \bottomrule
	\end{tabular}}
    \vspace{-0.4cm}
	\label{tab:sota2}
\end{table*}

\noindent \textbf{Evaluation Focused on Small Objects.}
We report mIoU on six small object categories of Cityscapes (i.e., pole, light, sign, rider, mbike, and bike) in Table \ref{tab:small}. Our ECGAN and ECGAN++ generate better mIoU than the state-of-the-art method (i.e., OASIS \cite{sushko2020you}) on all these small object classes.  We also show visualization results in Figure \ref{fig:small}, clearly confirming that the proposed method is highly capable of preserving small objects in the output.

\subsection{Ablation Study} 
\noindent \textbf{Baselines.} We conduct extensive ablation studies on three datasets to evaluate different components of the proposed method.
Our method has 7 baselines (i.e., B1, B2, B3, B4, B5, B6, B7) as shown in Table~\ref{tab:sota2}: 
B1 means only using the encoder $E$ and the proposed image generator $G_i$ to synthesize the targeted images;
B2 means adopting the proposed image generator $G_i$ and edge generator $G_e$ to produce both edge maps and images simultaneously;
B3 connects the image generator $G_i$ and the edge generator $G_e$ by using the proposed attention guided edge transfer module $G_t$;
B4 employs the proposed semantic preserving module $G_s$ to further improve the quality of the final results.
B5 uses the proposed label generator $G_l$ to produce the label from the generated image and then calculate the similarity loss between the generated label and the real one.
B6 uses the proposed pixel-wise contrastive learning and class-specific pixel generation methods to capture more semantic relations by explicitly exploring the structures of labeled pixels from multiple input semantic layouts.
B7 uses the proposed multi-scale contrastive learning method proposed in Eq. \eqref{eq:contrastive_multi} to learn more semantic relations from multi-scale features.
B8 is our full model and uses the proposed cross-scale contrastive learning method proposed in Eq. \eqref{eq:contrastive_cross} to learn more semantic relations from cross-scale features.
As shown in Table \ref{tab:sota2}, each proposed module improves the performance on all three metrics, validating the effectiveness.

\begin{figure}[!t] \small
	\centering
	\includegraphics[width=1\linewidth]{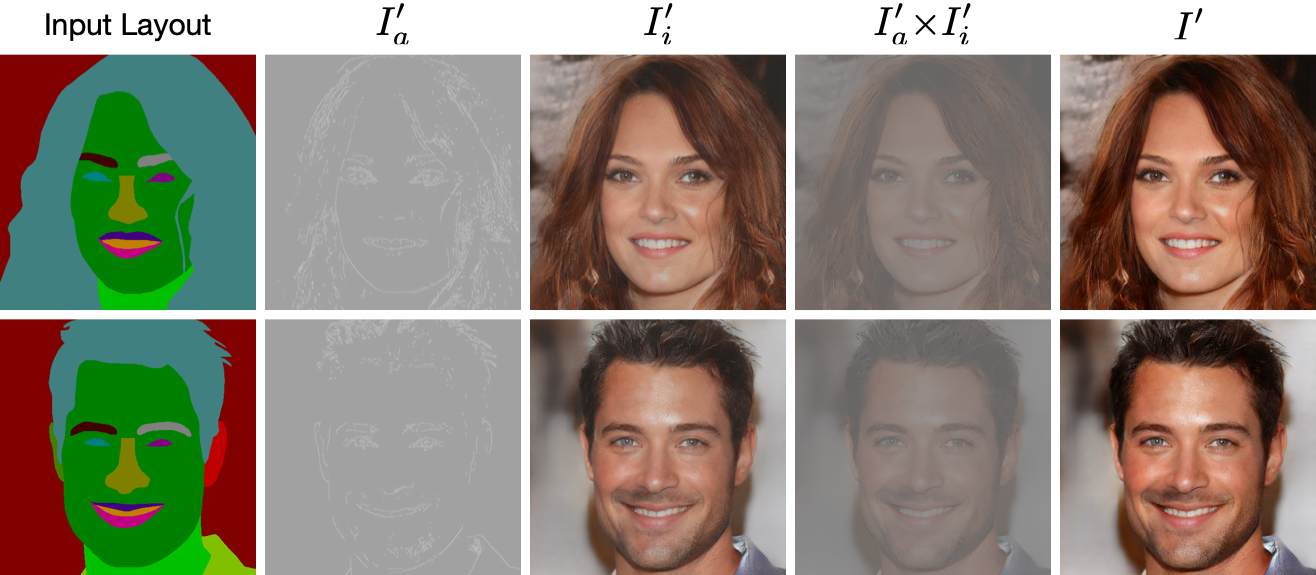}
	\caption{Visualization of the differences after the edge-guided refinement in Eq.~\eqref{eqn:image}.}
	\label{fig:diff2}
        \vspace{-0.4cm}
\end{figure}

\noindent \textbf{Effect of Edge Guided Generation Strategy.}
When using the edge generator $G_e$ to produce the corresponding edge map from the input label, performance on all evaluation metrics is improved.
We also provide several visualization results of the differences (see Eq.~\eqref{eqn:image}) after the edge-guided refinement in Figure~\ref{fig:diff2}.

\begin{figure*}[!t] \small
	\centering
	\includegraphics[width=0.9\linewidth]{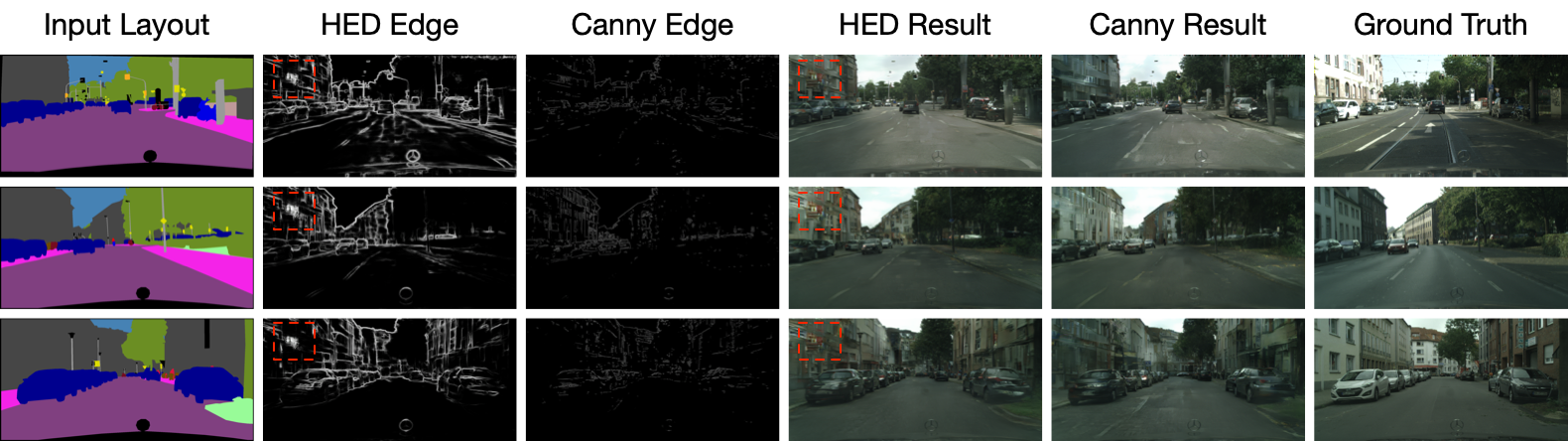}
	\caption{HED vs. Canny edge extraction.}
	\vspace{-0.4cm}
	\label{fig:edge}
\end{figure*}

\noindent \textbf{Effect of Edge Extraction Methods.}
We also conduct experiments on Cityscapes with HED \cite{xie2015holistically}, leading to the following results: 56.7 (FID), 64.5 (mIoU), and 82.3 (Acc), which are slightly worse than the results of Canny in Table \ref{tab:sota2}. 
The reason is that the edges from HED are very thick and cannot accurately represent the edge of objects. It also ignores some local details since it focuses on extracting the contours of objects. 
Thus, HED is unsuitable for our setting as we aim to generate more local details/structures.
Moreover, we see in Figure \ref{fig:edge} that the generated HED edges contain artifacts, as indicated in the red boxes, which makes the generated images tend to have blurred edges.

\noindent \textbf{Effect of Attention Guided Edge Transfer Module.}
We observe that the implicitly learned edge structure information by the ``$E$+$G_i$+$G_e$'' baseline is not enough for such a challenging task.
Thus we further adopt the transfer module $G_t$ to transfer useful edge structure information from the edge generation branch to the image generation branch.
We observe that performance gains are obtained on the mIoU, Acc, and FID metrics in all three datasets. 
This means that $G_t$ indeed learns rich feature representations with more convincing structure cues and details and then transfers them from the generator $G_e$ to the generator $G_i$.

\noindent \textbf{Effect of Semantic Preserving Module.}
By adding $G_s$, the overall performance is further boosted on all the three datasets.
This means $G_s$ indeed learns and highlights class-specific semantic feature maps, leading to better generation results.
In Figure~\ref{fig:city_seg}, we show some samples of the generated semantic maps. 
We observe that the semantic maps produced by the results with $G_s$ (i.e., ``Label by Ours II'' in Figure \ref{fig:city_seg}) are more accurate than those without using $G_s$ (``Label by Ours I'' in Figure \ref{fig:city_seg}). 
Moreover, we visualize three channels in~$\mathcal{F}^{'}$ on Cityscapes in Figure~\ref{fig:semantic_block}(right), i.e., road, car, and vegetation.
Each channel learns well the class-level deep representations.

\noindent \textbf{Effect of Similarity Loss.}
By adding the proposed label generator $G_l$ and similarity loss, the overall performance is further boosted on all three metrics. This means the proposed similarity loss indeed captures more intra-class and inter-class semantic dependencies, leading to better semantic layouts in the generated images.

\begin{figure}[!t] \small
	\centering
	\includegraphics[width=0.9\linewidth]{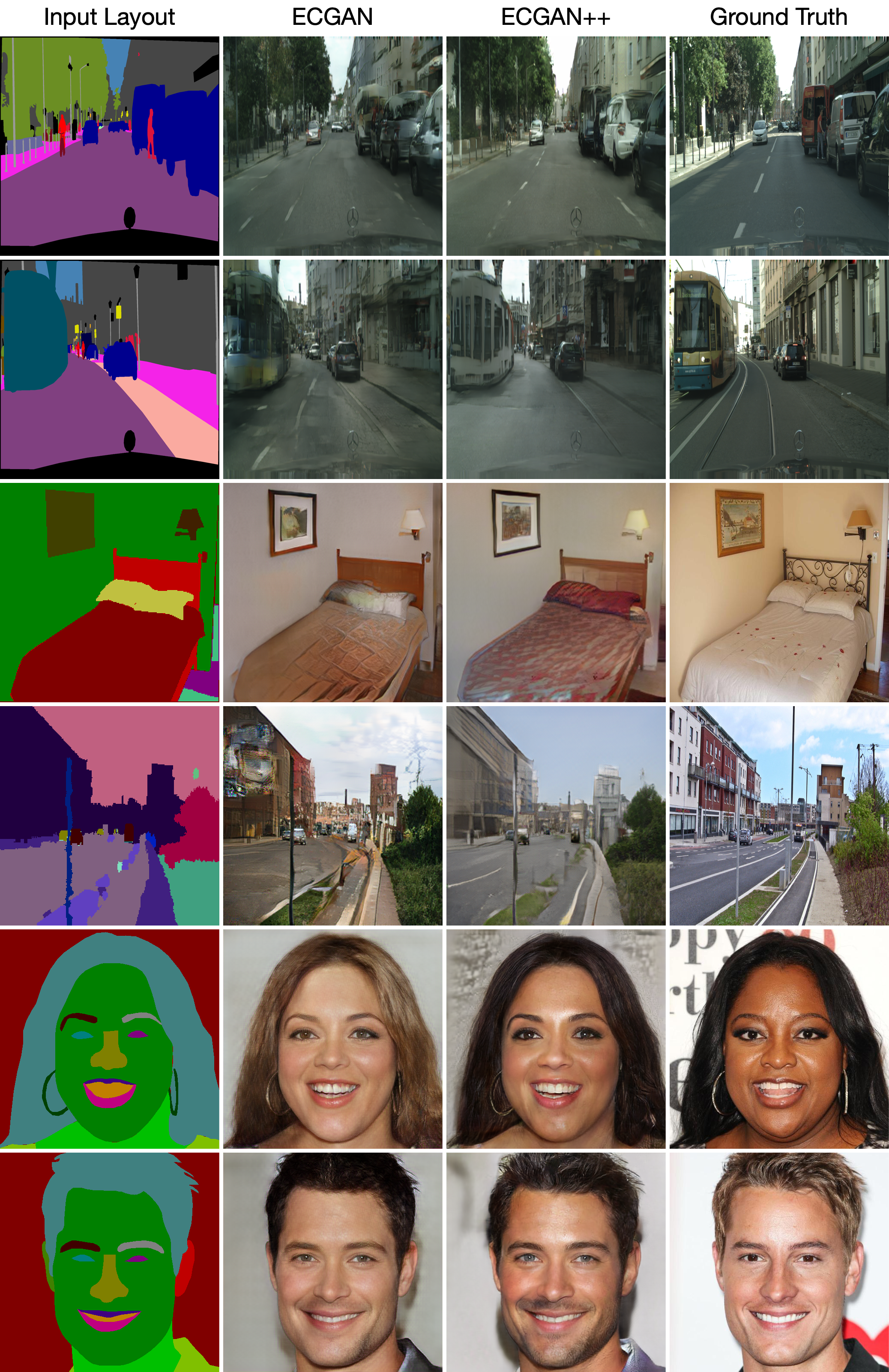}
	\caption{Comparison results of ECGAN and ECGAN++ on four datasets.}
	\label{fig:ecgan}
		\vspace{-0.4cm}
\end{figure}

\noindent \textbf{Effect of Contrastive Learning.}
When adopting the proposed pixel-wise contrastive learning module $G_c$ and class-specific pixel generation method to produce the results, the results are significantly improved on all three datasets on all three evaluation metrics.
This means that the model does indeed learn a more discriminative class-specific feature representation, confirming the superiority of our design.

\noindent \textbf{ECGAN vs. ECGAN++.}
We provide user study results in Table \ref{tab:atm2}. We can see that users favor the results generated by ECGAN++ on all three datasets compared
with those results generated by ECGAN.
We also provide quantitative comparison results of ECGAN~\cite{tang2023edge} and ECGAN++ in Tables~\ref{tab:sota} and \ref{tab:sota2}.
ECGAN++ achieves better results than ECGAN on all metrics on all the datasets.
Specifically, We see in Table \ref{tab:sota2} that B7 has better results than B6 on all datasets, and B8 has better results than B7 on the evaluation metrics, which verifies the effectiveness of our proposed multi-scale contrastive learning loss (Eq. \eqref{eq:contrastive_multi}) and cross-scale contrastive learning loss (Eq. \eqref{eq:contrastive_cross}).
Moreover, we note that ECGAN++ generates better results than ECGAN on the four datasets (including a face dataset CelebAMask-HQ \cite{CelebAMask-HQ}), as shown in Figure \ref{fig:ecgan}.
Finally, we compare the results of both ECGAN and ECGAN++ on the multi-model synthesis and small object generation evaluations in Table \ref{tab:multimodal} and \ref{tab:small}, respectively. We can see that ECGAN++ achieves much better results than ECGAN on both multi-model synthesis and small object generation evaluations, which verifies the effectiveness of the proposed multi-scale contrastive learning method.

\begin{table}[!t] \small
	\centering
 	\caption{Weight $w_s$ selection for the multi-scale contrastive learning loss in Eq. \eqref{eq:contrastive_multi}.}
% 	\resizebox{1\linewidth}{!}{% 
		\begin{tabular}{cccc} \toprule
 $w_s$ &  mIoU $\uparrow$ & Acc $\uparrow$ & FID $\downarrow$ \\ \midrule
           1.0 1.0 1.0 1.0 & 72.8 & 83.3 & 43.9 \\
           0.1 0.1 0.1 0.1 & 72.7 & 83.4 & 43.7 \\
           1.0 0.7 0.4 0.1 & \textbf{73.3} & \textbf{83.9} & \textbf{42.2} \\ \bottomrule
	\end{tabular}
    \vspace{-0.4cm}
	\label{tab:sele1}
\end{table}

\begin{table}[!t] \small
	\centering
 	\caption{Cross-scale pair selection for the cross-scale contrastive learning loss in Eq. \eqref{eq:contrastive_cross}.}
% 	\resizebox{1\linewidth}{!}{% 
		\begin{tabular}{ccccc} \toprule
Pairs & Setting &  mIoU $\uparrow$ & Acc $\uparrow$ & FID $\downarrow$ \\ \midrule
         0 & - & 72.8 & 83.5 & 43.7\\
         1 & (s4, s8) & 73.0 & 83.6 & 43.2 \\
         2 & (s4, s8), (s4, s16) & \textbf{73.3} & \textbf{83.9} & \textbf{42.2} \\ \bottomrule
	\end{tabular}
    \vspace{-0.4cm}
	\label{tab:sele2}
\end{table}

\noindent \textbf{Hyper-Parameter Selection.}
We also investigate the influence of
$w_s$ in Eq. \eqref{eq:contrastive_multi} on the performance of our model. The results of Cityscapes
are shown in Table \ref{tab:sele1}.
We see that the proposed method achieves the best results when applying a decreasing function (i.e., 1.0, 0.7, 0.4, 0.1) to the weights according to the output stride.
Moreover, we conduct ablation study experiments on the Cityscapes dataset to choose the number of cross-scale pairs in Eq. \eqref{eq:contrastive_cross}.
The results are shown in Table \ref{tab:sele2}, showing that two cross-scale pairs achieve the best results.
Considering the balance of training time and performance, we do not consider increasing the number of cross-scale pairs.

\section{Conclusion}

We propose a novel framework for semantic image synthesis.
It introduces four core components: edge guided image generation strategy, attention guided edge transfer module, semantic preserving module, and multi-scale contrastive learning module.
The first one is employed to generate edge maps from input semantic labels. 
The second one is used to selectively transfer the useful structure information from the edge branch to the image branch. 
The third one is adopted to alleviate the problem of spatial resolution losses caused by different operations in the deep nets.
The last one is utilized to investigate global-local semantic relations between training pixels from different scales, guiding pixel embeddings toward cross-image category-discriminative representations. 
Extensive experiments on three public datasets show that the proposed methods achieve significantly better results than existing models.

\hao{Although our method achieves good results on different datasets, our method also has a limitation, that is, it needs to be retrained on different datasets. In future work, we will design a new framework that can achieve good results on different datasets with only one training, which saves training time and training resources and is also more convenient to deploy to reality in the application.}

\noindent \textbf{Acknowledgments.}
This work was partly supported by the EU H2020 project AI4Media under Grant 951911.

%\clearpage
\small
\bibliographystyle{IEEEtran}
\bibliography{reference}

\begin{IEEEbiography}[{\includegraphics[width=1in,height=1.25in,clip,keepaspectratio]{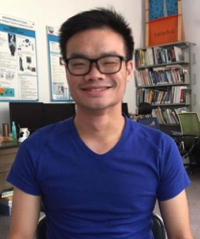}}]{Hao Tang}
is currently a Postdoctoral with Computer Vision Lab, ETH Zurich, Switzerland.
He received the master’s degree from the School of Electronics and Computer Engineering, Peking University, China and the Ph.D. degree from the Multimedia and Human Understanding Group, University of Trento, Italy.
He was a visiting scholar in the Department of Engineering Science at the University of Oxford. His research interests are deep learning, machine learning, and their applications to computer vision.
\end{IEEEbiography}

\begin{IEEEbiography}[{\includegraphics[width=1in,height=1.25in,clip,keepaspectratio]{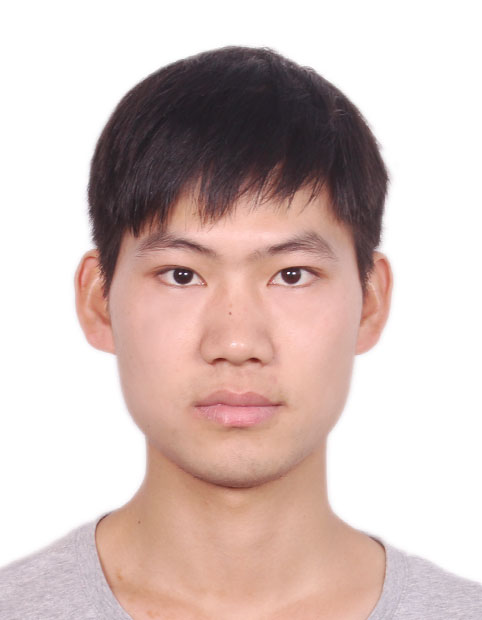}}]{Guolei Sun}
is a PhD candidate at ETHZ, Switzerland. He received master degree from King Abdullah University of Science and Technology in 2018. From 2018 to 2019, he worked as a research engineer at the Inception Institute of Artificial Intelligence, UAE. His research interests lie in computer vision and deep learning for tasks such as semantic segmentation, video understanding, and object counting. He has published about 20 top journal and conference papers such as TPAMI, CVPR, ICCV, and ECCV.
\end{IEEEbiography}

\begin{IEEEbiography}[{\includegraphics[width=1in,height=1.25in,clip,keepaspectratio]{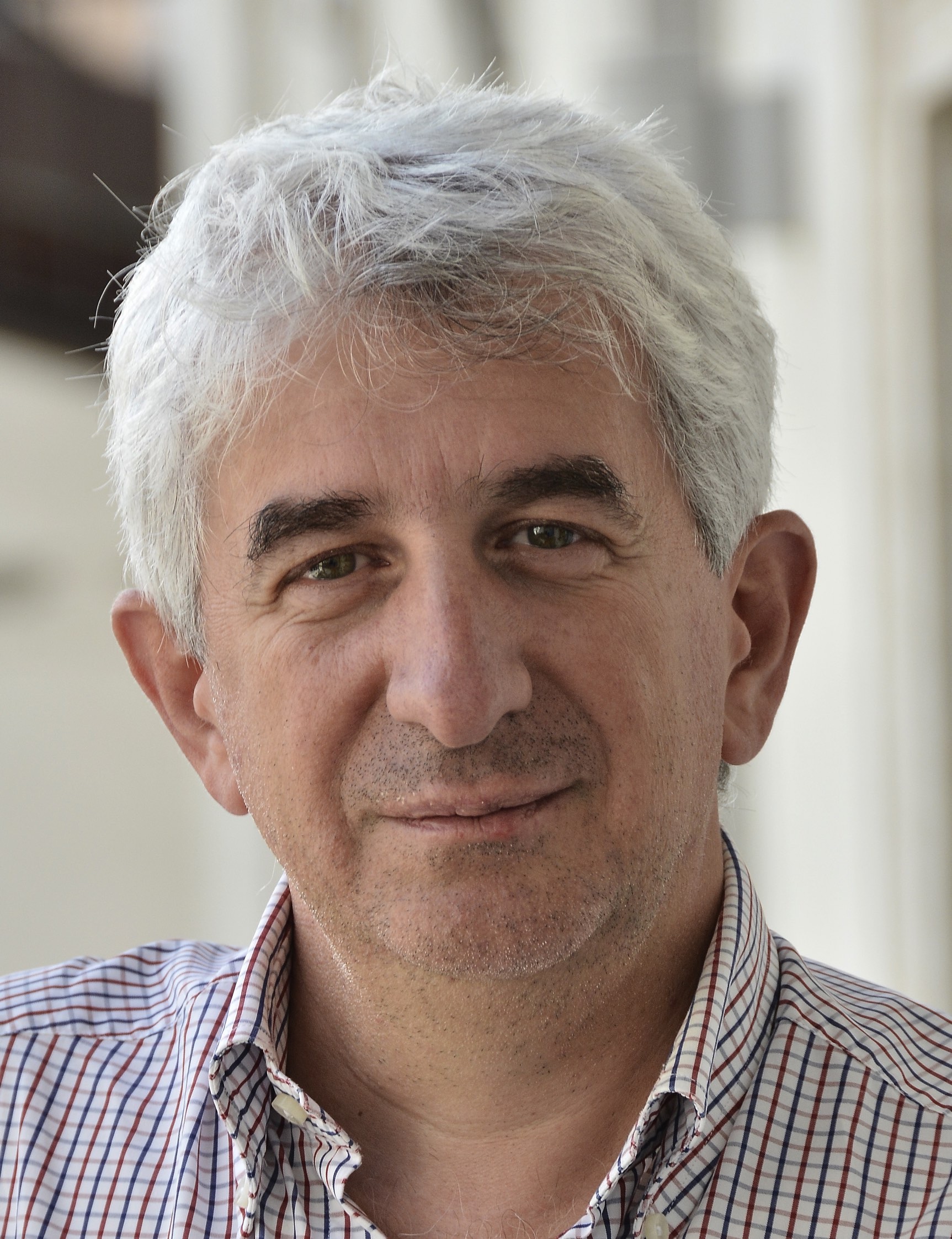}}]{Nicu Sebe} 
is Professor in the University of Trento, Italy, where he is leading the research in the areas of multimedia analysis and human behavior understanding. He was the General Co-Chair of the IEEE FG 2008 and ACM Multimedia 2013.  He was a program chair of ACM Multimedia 2011 and 2007, ECCV 2016, ICCV 2017 and ICPR 2020.  He is a general chair of ACM Multimedia 2022 and a program chair of ECCV 2024. He is a fellow of IAPR.
\end{IEEEbiography}

\begin{IEEEbiography}[{\includegraphics[width=1in,height=1.25in,clip,keepaspectratio]{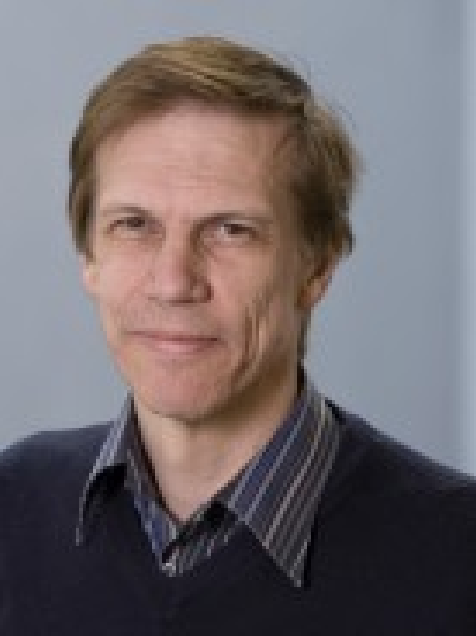}}]{Luc Van Gool} 
received the degree in electromechanical engineering at the Katholieke Universiteit
Leuven, in 1981. Currently, he is a professor at
the Katholieke Universiteit Leuven in Belgium and
the ETH Zurich in Switzerland. He leads computer vision research at both places, where he also
teaches computer vision. 
His main interests include 3D reconstruction and modeling, object recognition, tracking,
and gesture analysis, and the combination of those.
\end{IEEEbiography}


%\vfill


\end{document}